\documentclass[11pt]{article}
\usepackage[utf8]{inputenc}
\usepackage{lmodern}
\usepackage[T1]{fontenc}
\usepackage{amsmath, amsfonts, amssymb}
\usepackage{graphicx}
\usepackage{xr}
\usepackage[colorlinks=true, linkcolor=blue, citecolor=blue, urlcolor=blue]{hyperref}
\usepackage[numbers,sort&compress]{natbib}
\usepackage{geometry}
\usepackage{titlesec}
\usepackage{abstract}
\usepackage{authblk}
\usepackage{setspace}
\usepackage{booktabs}
\usepackage{array}
\usepackage[dvipsnames]{xcolor}
\titleformat{\section}
  {\normalfont\large}{\thesection}{1em}{\MakeUppercase}
\titleformat{\subsection}
  {\normalfont\normalsize}{\thesubsection}{1em}{\MakeUppercase}
\titleformat{\subsubsection}
  {\normalfont\normalsize}{\thesubsubsection}{1em}{\MakeUppercase}

\title{Towards generalizable deep ptychography neural networks}
\author[1]{Albert Vong}
\author[1]{Steven Henke}
\author[2]{Oliver Hoidn}
\author[1]{Hanna Ruth}
\author[1]{Junjing Deng}
\author[3]{Alexander Hexemer}
\author[3]{David Shapiro}
\author[2]{Apurva Mehta}
\author[2]{Arianna Gleason}
\author[4]{Levi Hancock}
\author[1]{Nicholas Schwarz\thanks{Corresponding author: nschwarz@anl.gov}}
\affil[1]{Argonne National Laboratory, Lemont, Illinois, USA}
\affil[2]{SLAC National Accelerator Laboratory, Menlo Park, California, USA}
\affil[3]{Lawrence Berkeley National Laboratory, Berkeley, California, USA}
\affil[4]{Brigham Young University, Department of Physics, Provo, Utah, USA}
\date{}

\newcommand{\resetsupplemental}{%
  \setcounter{section}{0}%
  \renewcommand{\thesection}{S\arabic{section}}%
  \setcounter{figure}{0}%
  \renewcommand{\thefigure}{S\arabic{figure}}%
  \setcounter{table}{0}%
  \renewcommand{\thetable}{S\arabic{table}}%
}

\begin{document}

\maketitle

\begin{abstract}
X-ray ptychography is a data-intensive imaging technique expected to become ubiquitous at next-generation light sources delivering many-fold increases in coherent flux. The need for real-time feedback under accelerated acquisition rates motivates surrogate reconstruction models like deep neural networks, which offer orders-of-magnitude speedup over conventional methods. However, existing deep learning approaches lack robustness across diverse experimental conditions. We propose an unsupervised training workflow emphasizing probe learning by combining experimentally-measured probes with synthetic, procedurally generated objects. This probe-centric approach enables a single physics-informed neural network to reconstruct unseen experiments across multiple beamlines—among the first demonstrations of multi-probe generalization. We find probe learning is equally important as in-distribution learning; models trained using this synthetic workflow achieve reconstruction fidelity comparable to those trained exclusively on experimental data, even when changing the type of synthetic training object. The proposed approach enables training of experiment-steering models that provide real-time feedback under dynamic experimental conditions.

\end{abstract}


\section{Introduction}

Ptychography has garnered significant interest over the past decade, particularly at large-scale X-ray facilities, including X-ray synchrotron light sources and X-ray Free Electron Lasers (XFELs). Its minimally destructive nature and exceptional spatial resolution have enabled breakthroughs in fields such as cellular imaging\cite{jiang2010quantitative}, nanoparticle engineering\cite{michelson2022three}, atomic-scale electron microscopy\cite{chen2021electron} and microelectronics characterization\cite{holler2017high}. Ptychography is classified as a Coherent Diffractive Imaging (CDI) technique and can exploit coherent diffraction to achieve high-resolution reconstruction beyond the physical limitations of traditional lens-based systems\cite{miao2025computational}.

A fundamental challenge in CDI is the phase retrieval problem: reconstructing the phase of diffracted light, which carries essential object information but is lost during detection due to intensity-only, i.e. squared amplitude, measurements\cite{shechtman2015phase}. This inverse problem is typically ill-posed; ptychography turns this tractable by introducing additional constraints through overlapping illumination measurements in real space or oversampling in reciprocal space\cite{gabor1948new, goodman2005introduction, faulkner2004movable, zheng2013wide}. These overlapping measurements enable simultaneous reconstruction of both the object and illuminating function, i.e. probe, using iterative algorithms\cite{thibault2008high}.

Current state-of-the-art iterative algorithms are computationally expensive\cite{maiden2009improved}. Next-generation light sources and detector upgrades have increased data acquisition rates by several orders of magnitude\cite{babu2023deep}\cite{aps_computing_strategy}, while reconstruction algorithms remain throughput-limited by their iterative nature. The resulting computational bottleneck is particularly problematic for experiment steering, where rapid feedback is essential for real-time decision making.

Deep neural networks (DNN) are an emerging alternative to conventional algorithms, potentially enabling real-time phase retrieval with reconstruction speed improvements spanning several orders of magnitude. Models for coherent diffractive imaging (CDI) reconstruction include convolutional neural networks, vision transformers, and measurement-guided diffusion models \cite{lee2025deep, vu2025pid3net, chang2023deep, wu2021complex, gan2024ptychodv, cherukara2020ai, deshpande2023investigating, nakahata2024ptychoformer, ye2022sisprnet, ding2022contransgan, guan2019ptychonet, pan2023efficient, cam2025ptychographic, wu2024fourier, hoidn2023physics}. Many models demonstrate some generalization on synthetic datasets and simple illumination schemes, e.g. plane-wave, relying on learned object priors to perform single-shot CDI reconstruction\cite{lee2025deep,nakahata2024ptychoformer, guan2019ptychonet, wu2024fourier, gan2024ptychodv}. However, successful DNN-based ptychographic reconstructions for real X-ray data remain scarce compared to demonstrations in visible light and transmission electron microscopy. This scarcity reflects exclusive domain-specific challenges: object diversity, non-trivial beam attenuation, and probe complexities like coherence, geometry and fluctuation. Current approaches are lacking in three main aspects: (1) not exploiting spatial overlap information inherent to ptychography measurements, reducing model accuracy; (2) implicitly conditioning on a single averaged probe function, which is not robust; (3) limited generalization studies across different experiment conditions.

In this work, we present a novel training workflow using ptychographic convolutional neural networks that addresses all three limitations identified above. By pairing experimentally-measured probes with synthetic objects for training, which emphasizes probe learning, we demonstrate robust synthetic-to-experimental domain transfer across diverse datasets spanning multiple instruments and facilities. We show that out-of-distribution performance depends critically on probe similarity between training and test conditions, reinforcing the importance of the probe. Using our synthetic workflow, our model generalizes across multiple dissimilar probes with minimal performance degradation. We envision our synthetic training strategy could enable deployment of generalizable models for experiment steering applications, where speed and robustness to changing conditions is valued over reconstruction quality.

\section {Results}
\label{sec: Results}

\subsection{Ptychography machine learning approaches}

The main goal of CDI techniques like ptychography is to reconstruct the phase image of an object, denoted $O$, given only a diffraction image $I$. The diffraction pattern originates from illuminating the object with a localized, coherent light source-the probe function $P$. For nanoscale imaging, achieving sub-100 nm spatial resolution requires probes focused by complex optics, resulting in spatially varying probe structures. The probe interacts with the object to create an exit wave $\psi = O \cdot P$. When measured in the far-field, the exit wave is observed as a unique diffraction pattern $I = |FT(\psi)|^2$ where all phase information is lost upon detection of the squared modulus. The task for conventional algorithms and DNN methods is to retrieve both the phase and the amplitude of $O$ and $P$, given only $I$-an ill-posed inverse problem that requires additional constraints for unique solutions. Ptychography addresses this challenge by acquiring multiple overlapping diffraction patterns containing shared information, thereby constraining the solution space and improving convergence guarantees \cite{dong2023phase}.

Conventional phase retrieval algorithms require numerous iterations to refine an object reconstruction, which can output high quality reconstructions but are computationally expensive. DNN-based methods seek to complement iteration-based reconstruction with surrogate models that directly map image (I) to object (O), yielding orders-of-magnitude time savings; see the following overview for alternative DNN approaches \cite{wang2024use}.

There are two general DNN approaches for phase retrieval. Feed-forward neural networks learn the mapping $G(O;I)$ during an initial training stage using large collections of image-object pairs, acquired either from experimental measurements with conventional reconstructions or from synthetic diffraction data generated using physics-based forward models. Once trained, these networks can infer on new, unseen data. In contrast, optimization-based neural methods (e.g. deep image priors, neural implicit representations) are more generalizable, but require time-intensive parameter optimization for each new dataset\cite{li2025learning, Zhou2023fpminr, bouchama2023physics, barutcu2022compressive, yang2022coherent, yang2021dynamic, wang2020phase}. We focus specifically on feed-forward methods due to their fast inference capabilities after initial training.

Current feed-forward DNN approaches face fundamental limitations in three main aspects: probe generalization, experiment robustness and CDI-based design. Regarding probe learning, most DNNs learn a conditional mapping $G(O ; I, \mathcal{P}_{train})$, where the training probe $\mathcal{P}_{\text{train}}$ defines a uniquely constrained mapping from diffraction to object if it lacks inversion symmetry\cite{chang2023deep}. Current methods condition on (1) a single non-trivial experiment probe \cite{chang2023deep, cherukara2020ai, babu2023deep, vu2025pid3net, yue2025physics, cam2025ptychographic}, (2) multiple similar synthetic probes \cite{nakahata2024ptychoformer, gan2024ptychodv, welker2022deep, bohra2023dynamic}, or (3) flat-field illumination\cite{sinha2017lensless, deng2020interplay, li2018spectral, ye2022sisprnet, wang2020phase, yang2022coherent, lee2025deep, wu2021complex}, which may not extend to probe-based ptychography. This makes the DNN susceptible to catastrophic failure if the test probe differs from training probe. DNN stability under varying experiment probe conditions, even minor, remains poorly-studied. Critically, to our knowledge, \textbf{whether DNNs can be conditioned on multiple unique experiment probes} is unexplored.

For experiment robustness, synthetic data training can increase object variety beyond what is available in experimental datasets. To our knowledge however, systematic assessment of synthetic-to-experiment domain transfer (different objects, probes, instruments) remains largely absent. This gap hinders understanding of model robustness across varying experiment conditions; even measurements from the same instrument exhibit domain shifts from sample variability and instrument drift that challenge model generalization.

Lastly, most ptychographic DNNs should more accurately be described as CDI DNNs, since they transform single diffraction image inputs to corresponding objects. This approach ignores adjacent overlapping images, which contain useful information fundamental to ptychography. A more principled approach would incorporate ptychography-specific inductive biases to fully exploit additional information from overlapping images. Previous work demonstrated benefits of spatial constraint maps $F_{c}$ that enforce real-space consistency in overlapping solutions\cite{hoidn2023physics}.

Incorporating other domain-specific inductive biases can further improve model robustness. For example, training DNNs to predict diffraction patterns rather than objects implicitly enforces diffraction physics\cite{hoidn2023physics,wang2024use}. This includes energy conservation in the Fourier transform and respecting real-space transformation symmetries inherent to diffraction imaging. 

The DNN used in this study builds on the ptychography deep learning framework PtychoPINN, which introduced physics-based inductive biases via: (1) a physics forward model using known probes to enable unsupervised learning, (2) enforcing spatial overlap consistency in the real space object image, i.e. overlap constraint, by simultaneously predicting on groups of diffraction patterns and (3) using a Poisson noise model to account for measurement stochasticity\cite{hoidn2023physics}. We employ an alternate implementation (PtychoPINN-torch) better suited for training on diverse datasets (see Methods for details, and SI Figure S1 for an inductive bias study). 

\begin{figure}[htbp]
    \centering
    \includegraphics[width=0.999\textwidth]{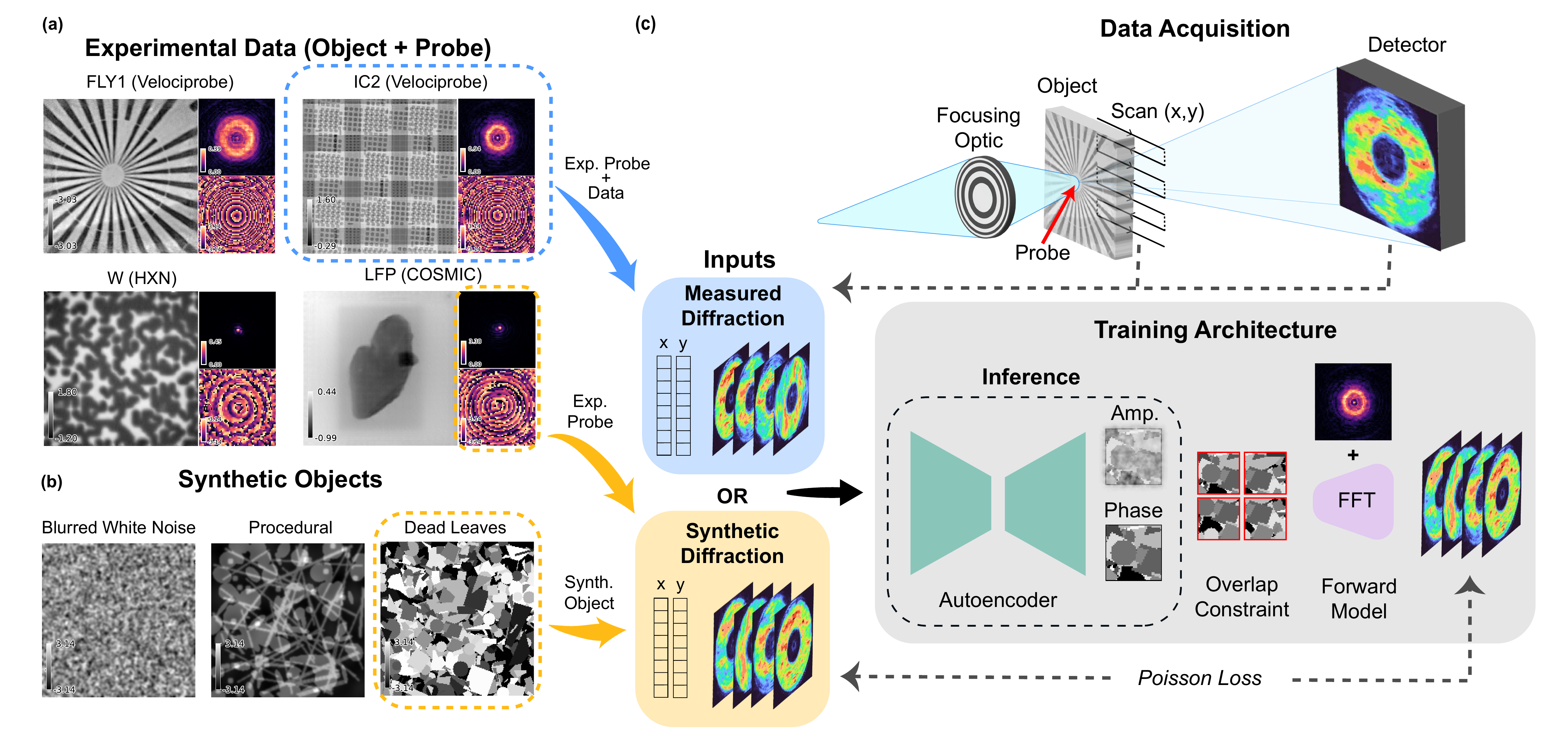}
    \caption{An overview of the data and PtychoPINN-torch training process. (a) Example experiment datasets, shown by their reconstructed objects (gray) and probes (pink). Note that only the phase structure in the center of the phase image associated with non-zero amplitude values (upper probe image) are valid.(b) Examples of synthetic object classes used to create synthetic training data. (c) The training process for PtychoPINN-torch. Training data is either sourced from experiment measurements or from synthetic data created from combining synthetic objects and reconstructed experiment probes. The data format containing diffraction images and probe positions is used directly for both training and inference. Training uses the full model -- the neural network combined with physics-based constraints -- while inference uses only the neural network mapping component.}
    \label{fig:dataset_showcase}
\end{figure}

We combine this robust DNN with a data approach combining the strengths of synthetic and experiment data:  we illuminate synthetic objects with experimentally-measured probes, which generates diverse training examples while grounding the neural network in realistic measurement conditions. The resulting synthetic diffraction patterns preserve the experimental probe's spectral characteristics and asymmetries, conditioning the DNN on experimental conditions seen at test time (see Methods for data generation details). We explore several classes of procedurally-generated objects, primarily using the Dead Leaves (DL) model due to its demonstrated effectiveness in CNNs\cite{baradad2021learning}\cite{yamada2024ptychographic}. Figure \ref{fig:dataset_showcase} shows the experimental and synthetic datasets alongside the training workflow, while Table \ref{tab:objects} describes synthetic objects used in this study.

\subsection{A diverse ptychography evaluation dataset}
\label{sec: experiments}

One important but previously unaddressed challenge in X-ray ptychography machine learning is the heterogeneity of experimental data formats across facilities and instruments, each with unique file structures and data conventions. This data heterogeneity makes it difficult to construct a diverse dataset to evaluate model robustness. We address this by curating a data corpus that represents diverse measurement setups at different ptychography beamlines. We collected experimental datasets from: Velociprobe (2-ID-D, Advanced Photon Source), Hard X-ray Nanoprobe (HXN, 26-ID, Advanced Photon Source - Center for Nanoscale Materials Sector 26), Cosmic Imaging (7.0.1.2, Advanced Light Source) and X-ray Pump Probe (Linac Coherent Light Source-II). Table \ref{tab:datasets} contains experimental descriptions of all datasets used. Datasets were standardized using Ptychodus and reconstructed using Pty-Chi, which are both software packages developed at the Advanced Photon Source (see Methods). We release this dataset alongside this publication to facilitate further studies on DNN generalization capabilities and cross-facility model development. These datasets comprise a variety of objects, measurement devices (Charge-Coupled and Photon-Counting Devices) and illumination characteristics from different light sources. Experiments from the same instrument are collected at different points in time, and contain differences in instrument alignment and illumination conditions.

We choose Fourier Ring Correlation (FRC) as our primary evaluation metric, which offers four key advantages: suitability for complex-valued predictions, relative insensitivity to scaling differences, detailed frequency-domain interpretation of reconstruction quality, and robustness in the absence of ground truth. Real-space, standard metrics such as Structural Similartiy Index Measure are inappropriate as PtychoPINN-torch is evaluated on datasets with no absolute ground truth and drastically differing measurement conditions\cite{fienup1997invariant}. Photon scaling factors are neither learned parameters nor internalized by the model, leading to non-trivial distribution shifts in amplitude and phase maps. To quantify the accuracy of out-of-sample predictions, we define FRC-AUC as the integral of the FRC curve from zero frequency to the 50\% threshold (Nyquist frequency), providing a single scalar performance measure \cite{cao2022automatic}.

\subsection{Single Experiment Transfer Learning}

\begin{figure}[htbp]
    \centering
    \includegraphics[width=0.999\textwidth]{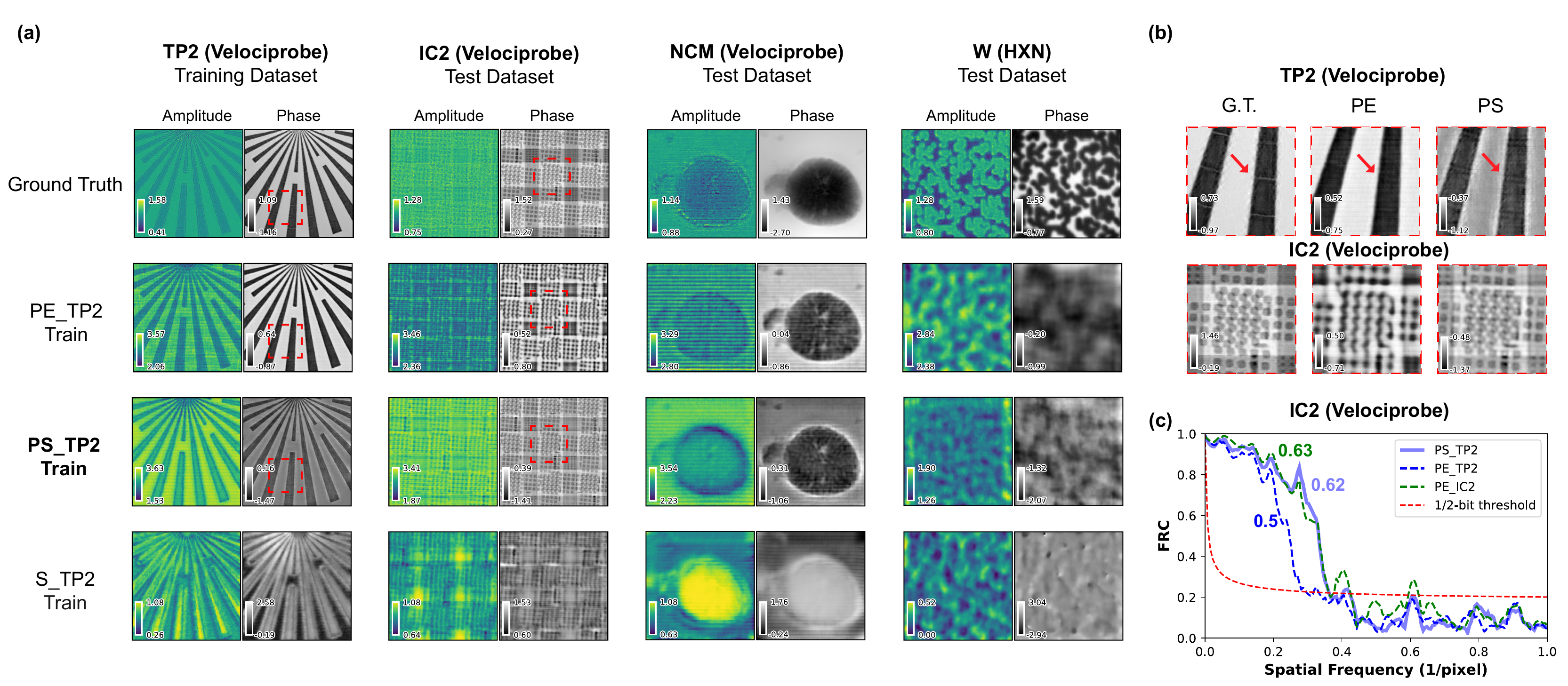}
    \caption{Comparison of models trained on synthetic (PS and SS) and experimental data (PE). a) From left to right: Reconstructions of datasets TP2, IC2, NCM and W from models trained using only TP2 raw data. PS\_TP2 exceeds or equals PE\_TP2 on test datasets from the same instrument (NCM, IC2), while a supervised model, SS\_TP2, transfers poorly across the board. All models generalize poorly on the W dataset, whose probe function differs greatly from the training probe function (due to being from a different instrument) and cannot be generalized to. b) Enlarged regions of the TP2 (top) and IC2 (bottom) datasets, with ground truth, PE and PS reconstructions. PE misses high frequency features in both training set (TP2) and test set (IC2). c) FRC comparison of 3 models on the IC2 dataset, with baseline PE\_IC2 (model trained exclusively on IC2 experimental data). PS\_TP2, despite being trained on a probe from a different experiment, performs nearly identically. PE\_TP2 performs the worst of the three, due to limited object variety in the TP2 dataset which does not generalize to the IC2 dataset.}
    \label{fig:exp_syn_comparison}
\end{figure}

We first compare models trained on synthetic data from single experiments (denoted \textit{PS\_name}) to baseline models trained only on experimental data (denoted \textit{PE\_name}), to investigate the effectiveness of our synthetic training approach. Here, we expect \textit{PE} models to establish the performance upper bound since they train directly on experimental measurement conditions and can overfit to dataset-specific characteristics. For comparison, we also evaluate an equivalent supervised architecture similar to previous studies\cite{hoidn2023physics,cherukara2020ai,yue2025physics}. We show reconstructions on three representative datasets from the Velociprobe and one dataset from the HXN instrument in Figure \ref{fig:exp_syn_comparison}: (1) TP2 is a large-scale test pattern featuring vertically-oriented spokes with cross-hatches. (2) IC2 is a large circuit-board sample with different degrees of phase contrast and feature length scales. (3) NCM is a catalyst nanoparticle with inner, high-contrast components. (4) W is a tungsten sample featuring round, connected segments.

Three models trained from experiment \textit{TP2} are shown in Figure \ref{fig:exp_syn_comparison}a: PE\_TP2 (trained on experimental diffraction patterns), PS\_TP2 (trained on synthetic diffraction patterns generated using the experimental probe from \textit{TP2}), and a baseline supervised model S\_TP2 (same training data as PS\_TP2 but in a supervised setting). We also show out-of-distribution predictions on test datasets \textit{IC2}, \textit{NCM} and \textit{W}. All other reconstructions, including the LCLS dataset, are available in the SI, Figures S2-11.

Several consistent patterns emerge across all reconstructions. First, PtychoPINN-torch overestimates amplitude quantities, which results from not enforcing amplitude bounds (see Methods); we find this trade-off improves phase generalization. Second, the DNN reconstructions from the Velociprobe instrument (\textit{IC2}, \textit{NCM}) where the training dataset was measured, are higher quality than from the Hard X-ray Nanoprobe (\textit{W}). This reflects how the learned mapping $G(O; I, P_{\text{train}})$ uniquely constrains the reconstruction space only for probe functions similar to $P_{\text{train}} = P_{\text{TP2}}$. As probes contain instrument-specific characteristics, the DNN cannot generalize to other instruments with differing probes, e.g. the Velociprobe-trained model does not generalize to the Hard X-ray Nanoprobe.

When reconstructing the training dataset \textit{TP2}, PE\_TP2 most accurately captures the phase distribution of the ground truth reconstruction, while PS\_TP2 exhibits reduced contrast between test pattern and background due to halo artifacts in both phase and amplitude reconstructions. The S\_TP2 reconstruction captures general structural features but demonstrates markedly inferior quality compared to both PtychoPINN-torch variants. This indicates that our synthetic training workflow alone is insufficient for generalization without a robust model.

Despite phase distribution shifts, PS\_TP2 surprisingly demonstrates superior high-frequency reconstructions across datasets. Figure \ref{fig:exp_syn_comparison}b shows a representative section of \textit{TP2} where PS\_TP2 reconstructs perpendicular hatch lines (red arrow) across the main vertical spokes that are nearly invisible in PE\_TP2 reconstructions. In dataset \textit{IC2}, PS\_TP2 captures amplitude/phase distributions more accurately than PE\_TP2, particularly for high-frequency features. This synthetic-to-experiment domain transfer demonstrates that synthetic object diversity can enable models to generalize across frequency bands underrepresented in experimental training data, while experimental probes ground the DNN to the probe-object distribution for a specific instrument.

The performance gap between PE and PS models depends considerably on the test dataset power spectral distribution. On dataset \textit{IC2} which contains high-spatial frequency features, PS\_TP2 achieves superior reconstruction quality (FRC-AUC of 0.56 vs. 0.51 for PE\_TP2), with clear visual improvements in the magnified reconstructions (Figure \ref{fig:exp_syn_comparison}b, bottom). Conversely, both models perform similarly on predominantly low-frequency datasets like \textit{NCM}. Remarkably, PS\_TP2 achieves comparable performance to models trained directly on individual, unseen datasets from the Velociprobe, despite only using the training probe from a different experiment. This demonstrates that the learned mapping $G(O;I,P_{\text{train}})$ can achieve reasonable reconstructions when train and test probes are similar but not identical. 

\subsection{Learning multi-probe representations}

Having shown that synthetic training data can be used to approximate the experimental mapping $G(O;I,P_{\text{train}})$ for a single probe, we investigate scaling to multiple training probes with fixed training data size. Under the assumption that each mapping must be independently learned, we expect performance degradation as the number of training probes increases: the model must learn additional mappings with less data per probe and fixed model capacity. We adopt the shortened notation $G(O;P)$ where diffraction intensity is implicit, and demonstrate our model can successfully learn multi-probe mappings using only synthetic training data.

We train PtychoPINN-torch on synthetic data generated from four dissimilar probe functions in datasets \textit{W}, \textit{FLY1}, \textit{IC2}  and \textit{LFP}; these span the three instruments Velociprobe, HXN and Cosmic. We evaluate four training scenarios with an increasing number of training probes (Figure \ref{fig:multi_probe_comparison}a) evaluated against a test dataset with probe $P_{\text{test}}$:

\begin{figure}[htbp]
    \centering
    \includegraphics[width=0.999\textwidth]{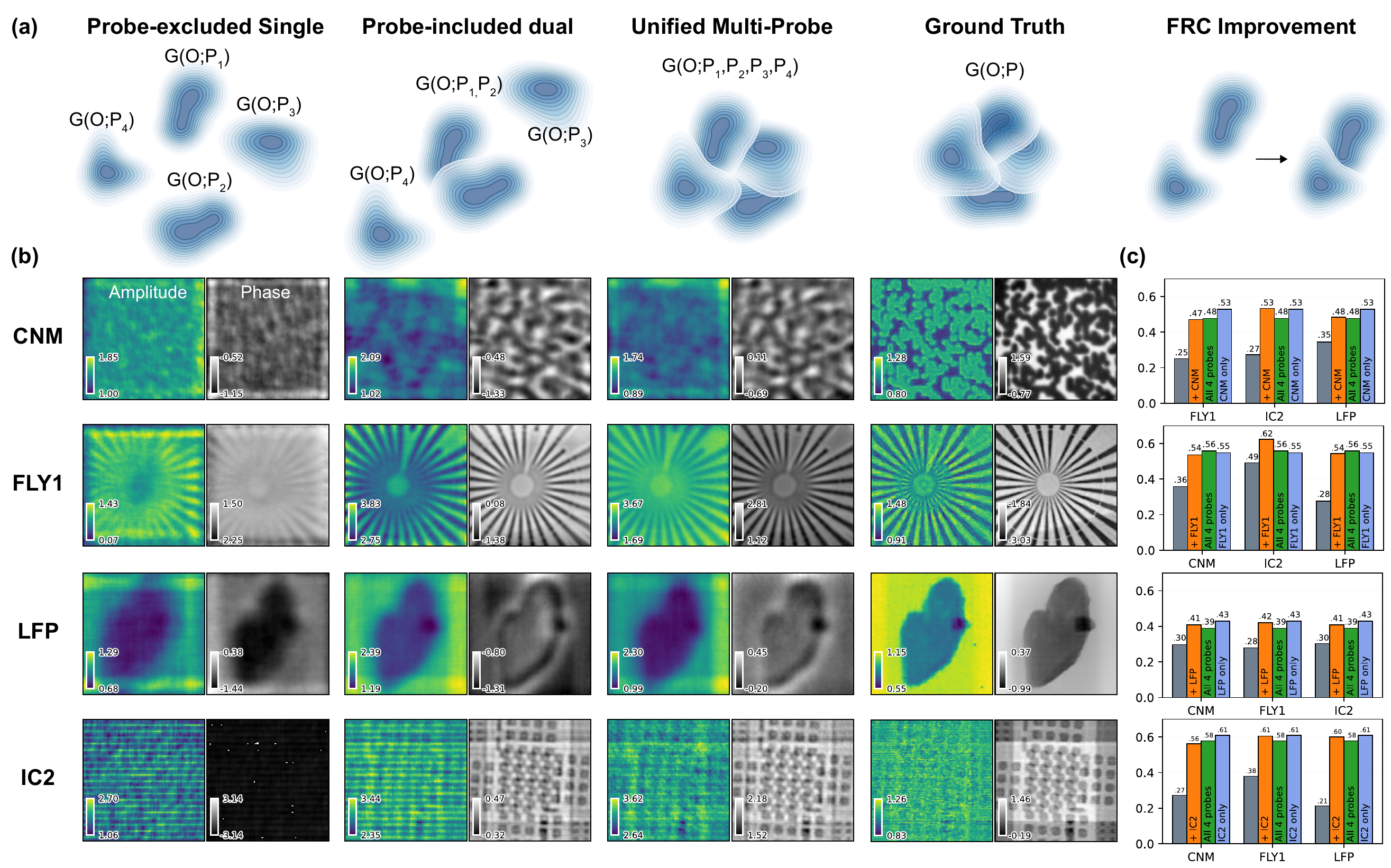}
    \caption{a) Schematic representation of probe-dependent conditional mappings for our training scenarios, where learned mappings for individual probes are largely independent of each other. As additional probes are added to the training dataset, the model is forced to learn a joint mapping that generalizes across all training probes. b) Reconstructions for datasets (instruments): \textit{W} (HXN), \textit{FLY1} (Velociprobe), \textit{IC2} (Velociprobe) and \textit{LFP} (Cosmic) under 3 training schemes with an increasing number of training probes from left to right. Under probe-excluded single training, models perform poorly when training and testing probe differ. When the testing probe is added to training in probe-included dual, the reconstruction quality remains high, up to 4 distinct training probes (unified multi-probe). c) FRC-AUC scores organized by experimental dataset per row (see b). Bars represent probe-excluded single training (gray), dual-probe training (orange), multi-probe training (green), and test probe-only training (blue). X-axis labels describe probes used in probe-excluded single training. }
    \label{fig:multi_probe_comparison}
\end{figure}

\begin{enumerate}
\item \textbf{Probe-excluded single training}: Models train on synthetic data from probe $P_i \neq P_{\text{test}}$, learning the mapping $G(O;P_{\text{i}})$. This tests whether unrelated probes provide useful inductive bias.

\item \textbf{Probe-included dual training}: Models learn the dual-probe mapping $G(O;P_i, P_{\text{test}})$ where the test probe is included in training alongside $P_i$ from probe-excluded single training. Synthetic dataset size remains constant and is split across both probes. 

\item \textbf{Unified multi-probe training}: A single model learns the comprehensive mapping \newline
$G(O;P_1, P_2, P_3, P_4)$ across all available probe functions. Synthetic data is split across all 4 probes.

\item \textbf{Test probe single training}: A baseline where the model uses its full capacity to learn the mapping for the a single test probe. Same model as probe-excluded single training, but evaluated on the training set.

\end{enumerate}

Probe-excluded single training yields poor reconstructions, as the learned mapping $G(O;P_{\text{train}})$ (Scenario 1) is strongly conditioned on the training probe (Figure \ref{fig:multi_probe_comparison}b, leftmost column). This leads to two complementary observations: (1) Cross-instrument inference fails because the learned mapping for a specific probe is too dissimilar to other probe mappings; (2) Similar probe characteristics, typically from same-instrument experiments, lead to more shared information in the learned mapping. For example, \textit{FLY1} and \textit{IC2} probes share partial phase structure despite having different amplitude rings (Figure \ref{fig:dataset_showcase}a), resulting in better relative FRC scores (\ref{fig:multi_probe_comparison}c, gray bars). As previously demonstrated, Velociprobe datasets such as \textit{NCM} and \textit{TP2} share both phase and amplitude probe structure, leading to high transfer learning accuracy (Figure \ref{fig:exp_syn_comparison}).

Probe-included dual training (Scenario 2), which incorporates the test probe alongside the poorly performing training probe, produces immediate quantitative and qualitative reconstruction improvements (\ref{fig:multi_probe_comparison}b, second column, and \ref{fig:multi_probe_comparison}c, orange bars). Surprisingly, dual-probe models perform similarly to the baseline single test-probe configuration (Scenario 4, \ref{fig:multi_probe_comparison}c, blue), while outperforming single-probe models trained exclusively on the same effective number of test probe training examples (14,000, see SI Figure S12). This performance gain is unexpected: dual-probe models must simultaneously learn separate mappings for the dissimilar probes P\_i and P\_test (14,000 images each). This suggests that dissimilar probe functions encode some shared structural information about the inverse mapping $G(O;P_1,P_2)$, which benefits joint learning with multiple training probes but is insufficient for probe-excluded training.

Building on this finding that dissimilar probes share transferable information, we extend training with all four distinct probes (Scenario 3, \ref{fig:multi_probe_comparison}b, third column and \ref{fig:multi_probe_comparison}c, green). This scenario shows minimal degradation in reconstruction quality compared to probe-included dual training, and more importantly, the baseline single test-probe (Scenario 4). This reveals PtychoPINN-torch's capacity to learn the joint representation $G(O;P_1, P_2, P_3, P_4)$ despite the relative uniqueness of individual mappings $G(O;P_i)$ and reduced per-probe training data. We discuss the implications of multi-probe training in the Discussion section.

\subsection{Frequency-dependent performance across synthetic objects}

\begin{figure}[htbp]
    \centering
    \includegraphics[width=0.999\textwidth]{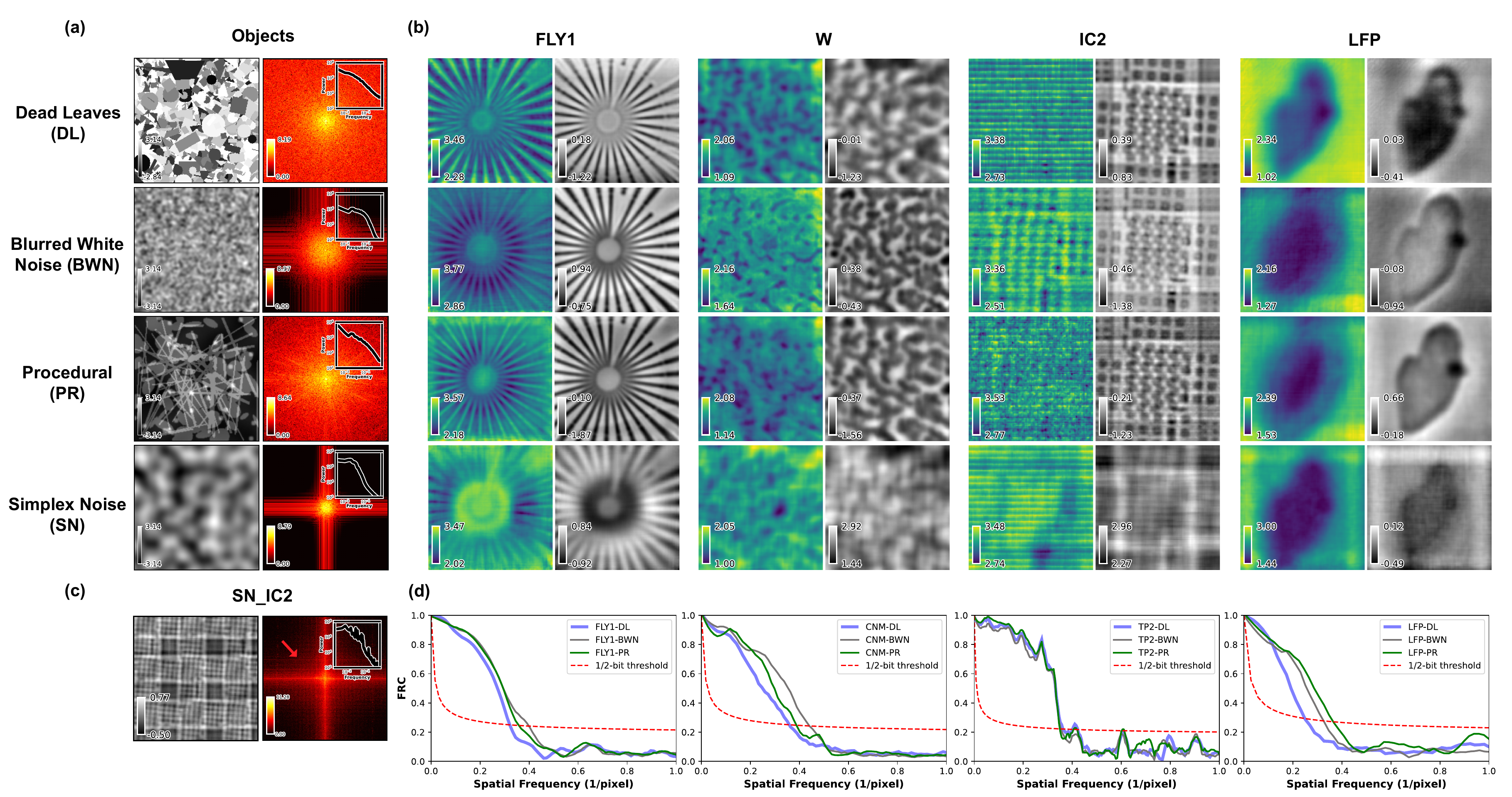}
    \caption{a) Images of synthetic objects dead leaves (DL), blurred white noise (BWN), procedural (PR) and simplex noise (SN). Each object image is accompanied by a 2-dimensional power spectral density (PSD) plot with an inset of the 1-dimensional integrated (PSD), showing differences in frequency statistics amongst synthetic images. b) Reconstruction results on datasets FLY1, W, IC2 and LFP for models trained exclusively on one synthetic object class. Each row corresponds to the synthetic class shown in a. c) Example reconstruction and PSD for simplex noise, showing high-frequency encoding via the probe rather than object. d) FRC curves of reconstructions from b showing differences in reconstruction quality reflecting the frequency statistics of the training datasets.}
    \label{fig:syn_object_descriptions}
\end{figure}

We demonstrate that different synthetic object classes with sufficient frequency diversity can train competitive models, with small performance differences reflecting the power spectral density (PSD) in their respective training datasets. We compare dead leaves (DL) with three alternative classes: geometrically-biased, semi-transparent polygons (PR), white noise blurred with a 3 pixel Gaussian kernel (BWN), and simplex noise (SN) (Figure \ref{fig:syn_object_descriptions}a). Two-dimensional PSD images (Figure \ref{fig:syn_object_descriptions}a, right) and radially integrated 1D PSD profiles (inset) show their distinct spectral characteristics: BWN consists of white noise convolved with a low-frequency Gaussian kernel, while SN exhibits low-frequency long-range correlations—both producing smooth features with uncovered frequency bands in the 2D PSD. PR provides broader frequency coverage, emphasizing thin straight lines and rounded edges (See SI Figure S13-S14 for model reconstructions of synthetic objects).

We reconstruct datasets using models trained exclusively on each object class (representative reconstructions in Figure \ref{fig:syn_object_descriptions}b), with naming convention Experiment-Object (e.g., IC2-DL denotes \textit{IC2} reconstruction with DL-trained model). IC2-DL, IC2-BWN and IC2-PR maintain consistent phase feature resolution, demonstrating that DL, BWN and PR contain adequate high-frequency components. Amplitude reconstructions have greater variability due to low sample absorption, which introduces noise in amplitude estimates (see \textit{IC2} ground truth reconstruction). Due to insufficient mid-to-high training frequencies, IC2-SN captures only coarse, periodic patterns.

Cross-dataset reconstructions reveal frequency-dependent trade-offs between training objects. BWN-trained models, having the largest proportion of low-to-mid frequency content, produce the best \textit{W} reconstruction, which shares similar spectral characteristics (Figure S15). Phase contrast in FLY1 is best captured by FLY1-DL, though FLY1-PR resolves the highest frequencies at center spokes, demonstrating complex trade-offs between phase distribution accuracy and resolution. LFP-DL best captures amplitude-phase correlation of the imaged nanoparticle, despite LFP-BWN and LFP-PR providing superior phase reconstructions.

SN-trained reconstructions exhibit higher-frequency features that equal or exceed those in the SN training data (Figure \ref{fig:syn_object_descriptions}c). IC2-SN contains grid points with a characteristic period of ~1/13 pixels$^{-1}$, matching the maximum frequency cutoff in SN. TP2-SN and NCM-SN also reconstruct features not represented in the SN dataset (see SI). This phenomenon arises from the coupled nature of the exit wave $\psi_{\text{exit}} = O \cdot P$, which encodes high-frequency information through the probe $P$. A DNN internalizes the probe frequencies through training, allowing it to reconstruct features beyond the training object's frequency bandwidth.

These results establish that performant ptychographic neural networks can be trained using diverse synthetic object classes, provided the probe function accurately represents experimental conditions. Our model's performance across different objects, combined with our demonstration that probe similarity between training and evaluation drives immediate performance improvements even for unseen experimental objects, reinforces the probe's central role in determining reconstruction quality. This asymmetry between object and probe importance suggests that ptychographic DNNs should prioritize investigating probe learning over object diversity to improve generalization capabilities.

\section{Discussion}

Our training workflow combines synthetic objects with experimental probes to achieve robust transfer learning from synthetic to experimental domains. Procedurally generated objects with multiple characteristic length scales provide adequate diversity to encode the conditional mapping $G(\cdot|P_{\text{measurement}})$, without requiring the use of traditional, natural image datasets. Our approach trains a functional mapping with only 28,000 training images for joint amplitude/phase reconstruction—substantially fewer than comparable studies. For example, Chang et al. required 250,000 ImageNet examples for electron ptychography phase-only retrieval for a smaller 32 x 32 probe, suggesting that PtychoPINN-torch's additional inductive biases substantially improve training efficiency and underlying representation learning\cite{chang2023deep}.

A key finding is PtychoPINN-torch's ability to jointly learn mappings for dissimilar probes with minimal reconstruction degradation,  using only a synthetic training workflow. We demonstrate robustness against varying photon scales spanning nearly two orders of magnitude (4.4 $\times 10^5$ to 9.6 $\times 10^6$). To our knowledge, this is among the first demonstrations of multi-probe learning in ptychographic DNNs, highlighting that this degree of information compression is even achievable by a lightweight architecture.

While our neural network reconstructions do not achieve the resolution or fidelity of conventional iterative algorithms, they provide sufficient quality for applications where rapid feedback is prioritized over accuracy. This performance trade-off enables DNNs to serve as tools for experiment steering rather than quantitative imaging replacements. During time-intensive tasks such as wide-area scans to locate regions-of-interest, replacing iterative reconstruction with single step inference dramatically improves measurement efficiency. For example, in large datasets like \textit{W}, we observe approximately 555x speedup compared to conventional reconstruction at 500 iterations (see Methods for details). These gains compound across repeated steering measurements, potentially accumulating time savings on the order of hours.

In our envisioned workflow, a DNN would be trained on multiple probes at the beginning of an experimental campaign, which can be acquired via previous or current experiments and optical simulations. Conditioning the DNN on a range of viable probes makes it more robust to changing experimental conditions such as shifting the sample focus distance. We have demonstrated that probes for a given instrument share substantial information, and anticipate that DNNs like PtychoPINN-torch can learn larger probe libraries than our demonstration using probes from different instruments. 

In conclusion, we introduced a novel training strategy for ptychographic neural networks that combines experimentally-grounded synthetic data with multi-probe learning, demonstrating robust performance on out-of-distribution experimental data. Our synthetic data training strategy, combined with a ptychography-specific neural network, can effectively learn inverse mappings $G(O;P_i)$ from diffraction patterns to object reconstructions. We show that model performance is strongly conditioned on the training probe distribution—large probe deviations during inference lead to reconstruction failure, emphasizing the importance of probe learning.  Our findings suggest a practical deployment strategy: training on libraries of realistic, instrument-specific probe variations enables robust experiment steering models. Multi-probe learning also opens promising research directions about probe capacity scaling in more expressive architectures such as vision transformers and diffusion models. Our approach significantly lowers the barrier for real-time feedback in X-ray ptychography experiments, where rapid qualitative feedback can substantially improve experimental efficiency and data quality.

\section{Methods}

\subsection{PtychoPINN-torch Implementation Details}

PtychoPINN-torch is based off the PtychoPINN architecture, which is a convolution neural network autoencoder \cite{hoidn2023physics}. The ptychographic overlap constraint requires the addition of a channel dimension in the input tensor, which represents different diffraction patterns which overlap in real space. This can result in duplication of the same diffraction image in multiple inputs, as a single diffraction image can belong to multiple overlapping "groups". We keep the same Poisson loss function as the original PtychoPINN model.

Differences from the PtychoPINN implementation include the replacement of the sigmoid activation function in the final layer of the decoder amplitude branch with a ReLu function. We found this replacement aided in out-of-distribution predictions due to the non-trivial shift in photon scales between training and testing datasets.

PtychoPINN-torch enhances several aspects of the PtychoPINN model and implementation. First, a custom dataloader was designed using the TensorDict framework in PyTorch. This allows for training on an arbitrary number of datasets, agnostic to photon scale and flexible to any number of measurement probes and scan patterns. TensorDict uses memory mapping, which enables \textit{scalable} training on large datasets that would otherwise exceed the memory constraints of modest graphics cards. This is particularly important for PtychoPINN, since each forward pass involves multiple images, image padding and translation operations, which all require additional memory. We note that this incurs a minor dataloading cost at the beginning of training and inference due to the instantiation of a memory map.

The second contribution of PtychoPINN-torch lies in the diffraction grouping algorithm. The current PtychoPINN implementation uses a KD-tree nearest neighbor search on scan coordinates to group nearby diffraction patterns, which randomizes the relative positions of images within the channel dimension. While this aggregation method guides the model toward a permutation-equivariant representation more robust to position jitter, it prevents the learning of inter-channel dependencies that arise from consistent spatial relationships between neighboring measurements.

Our approach instead maintains fixed position ordering to enable better inter-channel learning. For each reference point at origin (0,0), we systematically assign the 4 nearest neighbors to fixed quadrants of a 2D Cartesian grid: {0: (x < 0, y > 0), 1: (x > 0, y > 0), 2: (x < 0, y < 0), 3: (x > 0, y < 0)}. We define this type of grouping as "positional encoding". This fixed ordering allows the model to learn spatial correlations between pixels across channels, as neighboring diffraction patterns maintain consistent relative positions. Coordinate groupings are re-randomized during inference to ensure model robustness. Our implementation also handles experimental scan patterns where X and Y axes have dissimilar step sizes, which requires more careful handling of diffraction pattern grouping. It is also robust to non-cartesian scan paths such as spiral paths commonly used in ptychography measurements.

Third, we complement the fixed channel approach by incorporating Convolutional Block Attention Modules (CBAM) into the encoder. CBAM learns both spatial and inter-channel dependencies, which are consistent due to the constrained positioning of input channels. During training, we employ a two-stage fine-tuning strategy: first training the full network, then freezing the encoder (including CBAM modules) while allowing the decoder to refine object reconstruction using the feature representations learned by the enhanced encoder. We find that implementing CBAM reduces loss fluctuations when training on difficult synthetic datasets such as blurred white noise, while improving synthetic dataset generalization for objects such as blurred white noise.

\subsection{Coordinate grouping algorithm}

PtychoPINN-torch-specific diffraction pattern grouping is performed using a KD-tree search algorithm with range search, keeping the top n candidates. Distance-based filtering removes points outside a minimum and maximum range, allowing one to simulate different overlap conditions. These neighbors are then partitioned into one of four Cartesian quadrants relative to $\mathbf{r_i}$ which is fixed at the origin $(0,0)$. From each quadrant, one random neighbor candidate is sampled without replacement; $\mathbf{r_i}$ itself may be picked only once, for any quadrant. This yields spatially-ordered coordinate groups $\{\mathbf{r}_i^{(c)}, x_i^{(c)}\}_{c=0}^{3}$ forming training batches, where channel $c$ maps to quadrants: $\{0, 1, 2, 3\}$ defined above. Group-coordinate selection for a given $\mathbf{r}_i$ can also be repeated for additional sub-sampling during training or inference. This leads the input tensor to be of shape (\textit{Batch, Channel, Height, Width}), where Channel (C) represents spatially overlapping diffraction images from the same group.

\subsection{Training Details}

All training runs were conducted on 64 x 64 diffraction images, randomly split into training (95\%) and validation (5\%). We used 64 x 64 images instead of larger image sizes like 128 x 128, since reconstruction quality is sufficient at 64 x 64 and larger images result in slower inference speeds. We noticed no difference in generalization loss using a larger validation split, likely due to the unsupervised nature of training alongside the use of procedural objects for training. We therefore used a smaller validation set to increase the amount of training data for the model. Early stopping was used based on the validation loss, with the best performing model (i.e. validation) being saved and used for inference.

Model benchmarking is exclusively conducted on datasets unseen during training; either any other experimental dataset in the case of experiment-based training, and any experimental dataset in the case of synthetic-based training. Training used adaptive moment estimation and decoupled weight decay regularization (AdamW) \cite{loshchilov2017decoupled} as the optimizer. A universal learning weight of $10^{-3}$ was used, with a batch size of 16. Early stopping with a patience of 5 epochs was used based on the validation loss to save the best performing model checkpoint.

Experiment datasets use the data without additional augmentation for training, resulting in variable training dataset sizes. Synthetic datasets use a fixed number of 28,000 diffraction images, resulting from 4 unique objects with 7000 images each. The network trains on two NVIDIA GeForce RTX 4070 GPUs for 40 epochs, which takes 20 minutes for the synthetic dataset, and variable time for experimental datasets depending on diffraction image number.

\subsection{Supervised Model Implementation}

For comparison, we additionally implemented a supervised model keeping our autoencoder identical to ensure fair comparison. We apply a phase-centering procedure to all labeled phase images, where the phase image is split up into nine identical subsections (such as a tic tac toe board), and the mean phase value of the centered subsection is subtracted from all phase images as a pre-processing step\cite{babu2023deep}. The amplitude is unmodified from the ground truth reconstruction.

We apply the same RMSE photon scaling procedure to the inputs as our main PtychoPINN-torch model, and all of this is included as part of our custom dataloader under a "supervised" modality instead. For synthetic data, the object patches used to produce the diffraction patterns are included as labels, with the same phase subtraction procedure performed.

Our loss function takes a weighted sum of the MAE loss across both amplitude and phase images. In order to reflect the nature of our evaluated datasets, which have much more phase contrast than amplitude contrast, we apply a strongly weighted bias to the phase, with a ratio of 50:1. We found a large ratio helped the model prioritize the phase reconstruction details, instead of overly prioritizing the amplitude image which contains much less information. However, this approach still resulted in worse results than the PINN equivalent models (see SI).

\subsection{Evaluation Metrics}

As mentioned in the main text, we implement FRC inspired by the PtychoShelves library \cite{wakonig2020ptychoshelves}. Before the FRC itself is calculated, we perform a two-step sub-pixel registration, where fourier shifting is used based on phase alignment. Then, phase ramp is removed using a least squares fit. Finally, a soft-edge mask is applied to the object to remove edge effects for the FRC calculation. Code details can be seen in the accompanying code repository, in frc.py.

The FRC algorithm itself is standard, and uses the two-dimensional Fourier transforms of the two images we want to compare. See the following reference for additional details \cite{banterle2013fourier}.

\subsection{Data Preparation}

All experimental datasets are standardized and prepared using Ptychodus. Ptychodus is a software package from the Advanced Photon Source that acts as a common ptychography data pipeline. For a large number of supported instruments, it systematically standardizes disparate beamline formats, coordinate systems and scan metadata into a consistent, ML-ready data structure. It also has integrated capabilities with the Pytorch-optimized iterative package Pty-Chi, which allows for reconstruction of ground-truth data for experimental reconstruction verification.

All ground truth reference data in this manuscript was generated using Pty-Chi's least-squares maximum likelihood algorithm at 5000 iterations, with a single probe mode and no position refinement. The resulting object reconstruction (which we label as "ground truth" in the main text), alongside diffraction images and position data are packaged into a specific data format for training and inference. These can be found in the links to data.

Pre-processing is limited to removing saturated pixels by thresholding and flushing to zero, and physical to pixel coordinate conversion based on measurement geometry. Diffractions were cropped around a center pixel with either 64 or 128 pixel widths. The LCLS dataset specifically had additional pre-processing to remove diffraction patterns captured during source fluctuations.

Feature length scales vary across all datasets, from fine details in the IC2 dataset to long, straight edges in TP2 and FLY1. See SI Figure S1 contains 1D power spectral density (PSD) plots showing the distribution of frequencies in all images. The 1D power spectral density was obtained by doing a simple radial integration of the 2D image (see software repository for details). Notably, IC1 and IC2 have alot of frequency variation due to the presence of different-sized features within the images. On the other hand, NCM and LFP have the largest frequency drop-off, as they possess predominantly low-frequency features.

\subsection{PtychoPINN-torch diffraction pattern normalization}

Diffraction patterns are normalized on a per-dataset basis (i.e. scaled down) using a root-mean-square normalization,

\begin{equation}
n_{rms} = \sqrt{\frac{HW}{\frac{1}{N} \sum_{n=1}^{N} \sum_{i,j} I_n(i,j)^2}}
\label{eq: rms_norm}
\end{equation}

where N is the batch size, and i/j are row and column pixel indices. This normalization
amplifies high-intensity diffraction features while suppressing low-signal regions, effectively prioritizing the most informative parts of each pattern for reconstruction. The same scale factor rescales the unit-normalized network output back to experimental photon scales.

We also tried normalizing input diffraction patterns using an mean intensity normalization:

\begin{equation}
n_{energy} = \frac{1}{\frac{1}{N} \sum_{n=1}^{N} \sum_{i,j} I_n(i,j)}
\label{eq: rms_energy}
\end{equation}

While this approach enforces physical energy conservation, it yielded much worse reconstruction quality in practice, likely due to inadequate emphasis on high-information diffraction regions.

Probe functions are standardized using root-mean-square normalization, rather than energy-preserving normalization. We observe a trade-off in effectiveness depending on training data composition: energy normalization degrades reconstruction quality when training on individual experimental datasets, but improves performance when training datasets containing a range of photon scales (e.g. multiple experiments). This suggests that probe normalization helps the network learn scale-invariant features, but may constrain learning when probe variations are minimal (i.e. single experiments).

\subsection{Synthetic Diffraction Data Generation}

The dead leaves model was generated using existing code \cite{baradad2021learning}. Other generation algorithms can be found in the accompanying software repository. These objects are converted to correlated amplitude-phase pairs by assigning randomized refractive index values to procedurally-placed features. The refractive index ranges were chosen to simulate typical experimental conditions: weak absorption with strong phase contrast. Amplitude values are constrained to [0.7, 1.0], while phase values are rescaled to [-$\pi$, $\pi$]. This phase scaling approach assigns an arbitrary phase center based on its specific object content, mimicking the arbitrary phase offsets encountered in experimental reconstructions. We observe some domain shift in the phase/amplitude distributions of the synthetic data versus the experimental data; synthetically-trained model predictions often do not predict phase/amplitude contrast in experimental datasets as accurately as models trained on experiment data.

64 x 64 synthetic diffraction patterns were generated using the measurement model 

\begin{equation}
x_{syn}(\mathbf{r}) = |FT(y_{syn}(\mathbf{r}) \cdot p_{exp}(\mathbf{r}))|^2
\label{eq: ft_obj}
\end{equation}
, where synthetic object patches $y_{syn}$ are multiplied by \textbf{experimental probe functions} $p_{exp}$ to simulate realistic measurement conditions. \textit{syn} and \textit{exp} denote synthetic and experimental origin, respectively. The resulting diffraction patterns $x_{syn}$ for each unique object $y_{syn}$ are illuminated with a Poisson photon distribution with a mean in [$10^4$, $10^6$], following the range of experiment photon scales (see Table \ref{tab:datasets}). This scaling is applied independently of probe choice, enabling intensity diversity even within datasets generated from a single probe function. This mimics multiple experiments performed at a single beamline,  where probe characteristics remain similar but photon flux varies between measurements.

Measurement positions $\mathbf{r}_i$ follow randomized experimental scan patterns: \textit{isotropic} (similar x,y steps) or \textit{rectangular} (asymmetric axis step). Object patches with sub-pixel interpolation are extracted via PyTorch's \textit{grid\_sample} function before applying the measurement model.

See SI figure S13 and SI figure S14 for additional reconstructions from simplex noise, as well as reconstructions of the training datasets themselves. All evaluations of synthetic models in the main text are on test sets of objects (i.e. the object datasets were generated at evaluation time and not seen or trained on beforehand).

\section{Author Contributions}.

A.V., S.H., O.H., N.S., A.M., A.H., conceived and designed the computational experiments. A.G., L.H., J.D. collected the experimental data. O.H., A.G., L.H., S.H., J.D. analyzed and prepared the raw data. A.V., O.H., designed and programmed the neural networks. A.V. carried out all computational experiments. A.V., S.H., O.H., J.D., N.S., wrote the paper. All authors reviewed and edited the paper.

\section*{Acknowledgments}

This work is supported by the U.S. Department of Energy (DOE) Office of Science-Basic Energy Sciences awards Collaborative Machine Learning Platform for Scientific 
Discovery and Collaborative Machine Learning Platform for Scientific Discovery 2.0. This work also received support from the DOE Office of Science ASCR Leadership Computing Challenge (ALCC) through the 2025–2026 award Enhancing APS-Enabled Research through Integrated Research Infrastructure. This research used resources of the Advanced Photon Source (APS) and the Argonne Leadership Computing Facility (ALCF), both U.S. DOE Office of Science user facilities operated by Argonne National Laboratory under Contract No. DE-AC02-06CH11357. This research used resources of the Advanced Light Source (ALS), a U.S. DOE Office of Science user facility operated by Lawrence Berkeley National Laboratory under Contract No. DE-AC02-05CH11231. This research used resources of the Linac Coherent Light Source (LCLS), a U.S. DOE Office of Science user facility operated by SLAC National Accelerator Laboratory under Contract No. DE-AC02-76SF00515.

The U.S. Government retains for itself, and others acting on its behalf, a paid-
up nonexclusive, irrevocable worldwide license in said article to reproduce, prepare derivative works, distribute copies to the public, and perform publicly and display publicly, by or on behalf of the Government. The Department of Energy will provide public access to these results of federally sponsored research in accordance with the DOE Public Access Plan. http://energy.gov/downloads/doe-public-access-plan.

\section*{Data availability}

All experimental data and reproducing model artifacts can be found at: \url{https://doi.org/10.5281/zenodo.16968020}. 

\section*{Conflict of Interest Statement}

The authors declare no competing interests.

\section{Code availability}

The python implementation for PtychoPINN-torch (alongside reproducibility notebooks) can be found at: \url{https://github.com/AdvancedPhotonSource/PtychoPINN-torch-pub}


\newpage

\begin{table}[htbp]
\centering
\vspace{12pt}
\begin{tabular}{@{}ll@{}}
\toprule
\textbf{Name} & \textbf{General Description} \\
\midrule
\textit{Procedural (PR)} &  Procedural lines/ellipses with empty space\\
\textit{Dead Leaves (DL)} &  Procedural shapes completely filling canvas\\
\textit{White Noise (WN)} & Single pixel values drawn from $N(0,1)$\\
\textit{Blurred White Noise (BWN)} & White noise blurred with Gaussian kernel\\
\textit{Simplex Noise (SN)} &  Low frequency noise\\

\bottomrule
\end{tabular}
\caption{Objects used for synthetic diffraction data. All object-generating functions are procedural or statistical in nature, leading to a variety of spectral characteristics.}
\label{tab:objects}
\end{table}

\begin{table}[htbp]
\centering
\vspace{12pt}
\begin{tabular}{@{}cccccc@{}}
\toprule
\textbf{Name} & \textbf{Instrument/Source} & \textbf{Image \# } & \textbf{\shortstack{Photons\\per image}} & \textbf{General Description} \\
\midrule
\textit{FLY1} & Velociprobe (APS) & 10,304 & $4.4 \times 10^{5}$ & Pattern w/ background features\\
\textit{TP1} & Velociprobe (APS) & 1,443 & $1.8 \times 10^{6}$ & Pattern w/ no background \\
\textit{TP2} & Velociprobe (APS) & 7,709 & $9.8 \times 10^{5}$ & Pattern w/ alignment markers \\
\textit{IC1} & Velociprobe (APS) & 1,443 & $9.4 \times 10^{5}$ & Zoomed-in circuit board \\
\textit{IC2} & Velociprobe (APS) & 9,436 & $9.4 \times 10^{5}$ & Large circuit board \\
\textit{NCM} & Velociprobe (APS) & 2,466 & $2.5 \times 10^{6}$ & LiNiCoMn$O_{2}$ particles \\
\textit{W} & HXN (APS-CNM) & 25,921 & $9.6 \times 10^{6}$ & Tungsten test pattern\cite{babu2023deep} \\
\textit{LFP} & Cosmic (ALS) & 5,625 & N/A & \parbox[t]{5cm}{\centering Catalyst particle\cite{marcus2021ptychography}\\Acquired on CCD detector} \\
\textit{TP-LCLS} & XPP (LCLS) & 1,572 & $1.2 \times 10^{7}$ & Pattern w/ background \\
\bottomrule
\end{tabular}
\caption{Dataset details for all experiments used. Includes illumination conditions, number of images and general description.}
\label{tab:datasets}
\end{table}

\bibliographystyle{naturemag}  
\bibliography{main_references}

\clearpage


\resetsupplemental  

\begin{center}
{\large \textbf{SUPPLEMENTAL INFORMATION}}

\vspace{8pt}

{\normalsize \textbf{Towards generalizable deep ptychography neural networks}}

\vspace{4pt}

{\small Albert Vong, Steven Henke, Oliver Hoidn, Hanna Ruth, Junjing Deng, Alexander Hexemer, Arianna Gleason, Levi Hancock, Apurva Mehta, Nicholas Schwarz}
\end{center}

\vspace{12pt}

\section{Investigating PtychoPINN-torch's inductive biases}
\label{inductive_biases}

The synthetic dataset advantage depends critically on PtychoPINN-torch's inductive biases, which constrain the model to learn physically-consistent mappings. We compare PtychoPINN-torch with three ablated variants with some or all of these inductive bias components removed: a supervised version without overlap constraints nor the forward mapping $F_{d}$, a PINN model with forward mapping $F_{d}$ but no overlap constraints, and a PtychoPINN model without positional encoding (coordinates are arbitrarily permuted in the input channels, which enforces coordinate permutation equivariance). Besides the removal of these components, the autoencoder architecture (which includes CBAM) remains constant between all models to provide a fair comparison. RMSE scaling is also applied to inputs for all models, \textit{including} the supervised model, which allows us to evaluate out-of-distribution performance. All models were trained on synthetic datasets except for the supervised model, which had to be trained on an experimental dataset to get reasonable reconstructions. The datasets selected highlight strengths of the different inductive biases, especially for high frequency features.

\begin{figure}[htbp]
    \centering
    \includegraphics[width=0.999\textwidth]{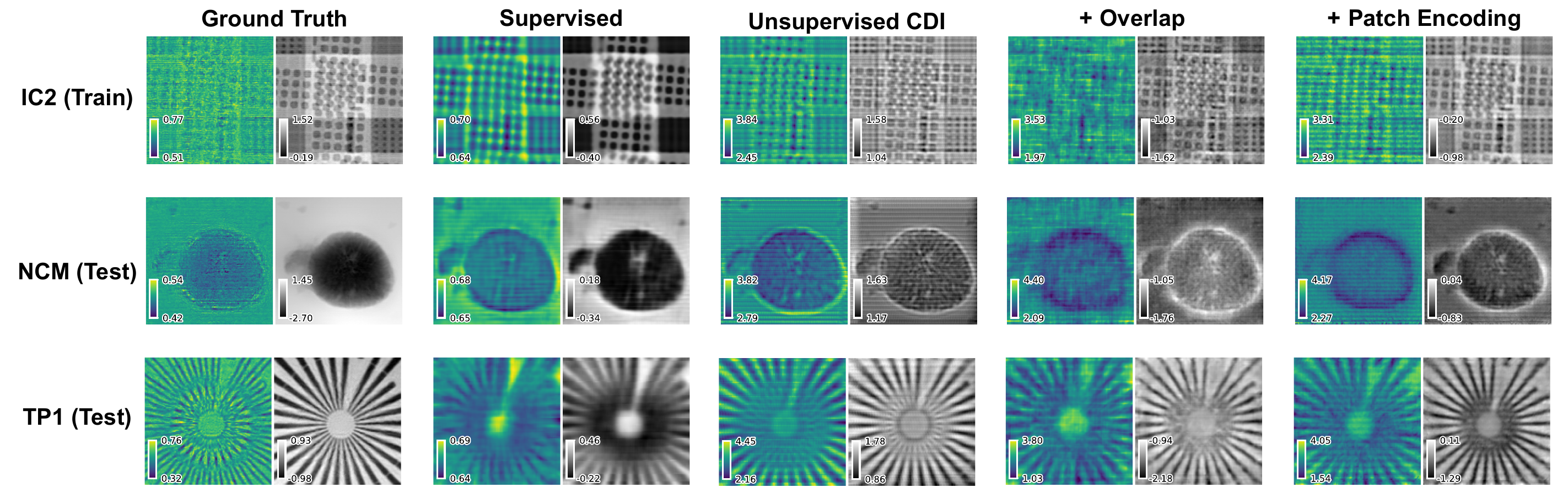}
    \caption{Ablation study results of PtychoPINN-torch's inductive biases. Each row represents a training or testing dataset trained with the following models from left to right: supervised, PINN, vanilla PtychoPINN and PtychPINNv2.}
    \label{fig:ablation_study}
\end{figure}

The supervised model has the worst reconstructions across all experiments including the training set, demonstrating poor generalization to different objects and measurement modalities. In \textit{IC2}, it overlearns the amplitude in the training dataset despite the bias weighting. It also exaggerates the contrast between regions, and misses much of the fine details within each circuit board section. It also does not fare well in a transfer learning setting despite seeing minimal probe drift. Both \textit{NC}M and \textit{TP1} show extremely degraded reconstructions where most high frequency details are omitted. It most accurately estimates the background-to-foreground contrast for \textit{NCM}, demonstrating that the synthetic training data likely imposes additional data biases to models that does not perfectly represent the test datasets.

The PINN model fares much better by internalizing diffraction physics in its learned mapping. However, the lack of overlap constraint introduces \textit{centrosymmetric ambiguity}, producing two object solutions for a given diffraction pattern. The stitching process averages ambiguous solutions, leading to the prominence of scan line patterns across the \textit{IC2} and \textit{NCM} datasets. Additionally, the PINN model lacks channel sharing as it learns using a CDI approach (i.e. one input diffraction image to one output object image), which mutes image contrast, as that typically requires more global context. Surprisingly, we found that the standalone PINN model excels specifically at flywheel reconstruction seen in \textit{TP1}, being able to finely reconstruct the center spoke pattern. We attribute this to position jitter which can obfuscate the predictions when channel-sharing is permitted (i.e. PtychoPINN), in addition to a low scan number, which reduces the signal to noise for the PtychoPINN-torch reconstruction.

The vanilla PtychoPINN model with overlap constraints generally fares worse than the PINN model in NCM and TP1. In IC2, where high frequency details are very important, channel sharing helps mitigate the centrosymmetric ambiguity, showing smoother reconstructed patches than the PINN model, with minimal scan-line patterns. However, there are "splotchy" regions throughout the reconstruction, showing a lack of consistent phase prediction for objects in the same region.

Finally, Ptychopinn-torch with patch encoding fares with \textit{IC2} and \textit{NCM}, but worse with \textit{TP1} versus the PINN model. In \textit{IC2}, there is phase region consistency with minimal scan line patterns. The \textit{NCM} shows the best phase contrast between inner nanoparticle components, despite the general phase-distribution shift in the background and foreground.

Notably, all three PINN-based models show less amplitude overlearning especially with \textit{IC2}. While we had to set specific ratios between the MAE loss components for amplitude and phase for a supervised learning approach, the forward model forces the DNN to automatically prioritize relevant details to reconstruct the original signal, which is largely phase information in the case of \textit{IC2}. Notably, this does not require manual tuning of relative importance weights for amplitude and phase. 

\section{Single dataset transfer learning reconstructions}

\begin{figure}[htbp]
    \centering
    \includegraphics[width=0.999\textwidth]{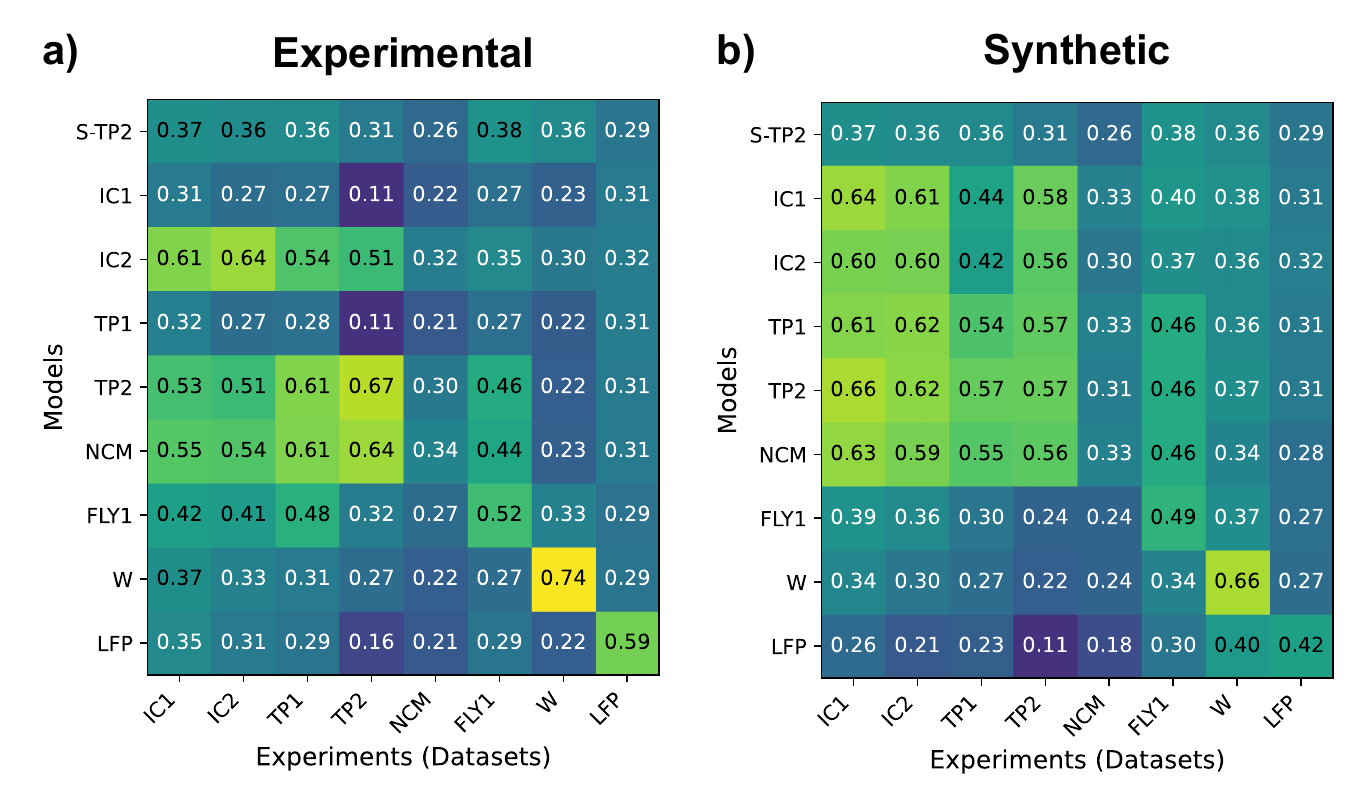}
    \caption{FRC-AUC comparison across experiments. Y-axis axis labels represent models trained on a single experimentla dataset, while x-axis labels correspond to dataset. a) Results for models trained on experimental data only b) Results for models trained on synthetic data only. Note that many predictions on the velociprobe datasets (\textit{IC1} to \textit{FLY1} from left to right) are fairly homogeneous for the synthetic model, as the probe functions are highly similar and have high predictive power.}
    \label{fig:single_dataset_summary}
\end{figure}

\begin{figure}[htbp]
    \centering
    \includegraphics[width=0.999\textwidth]{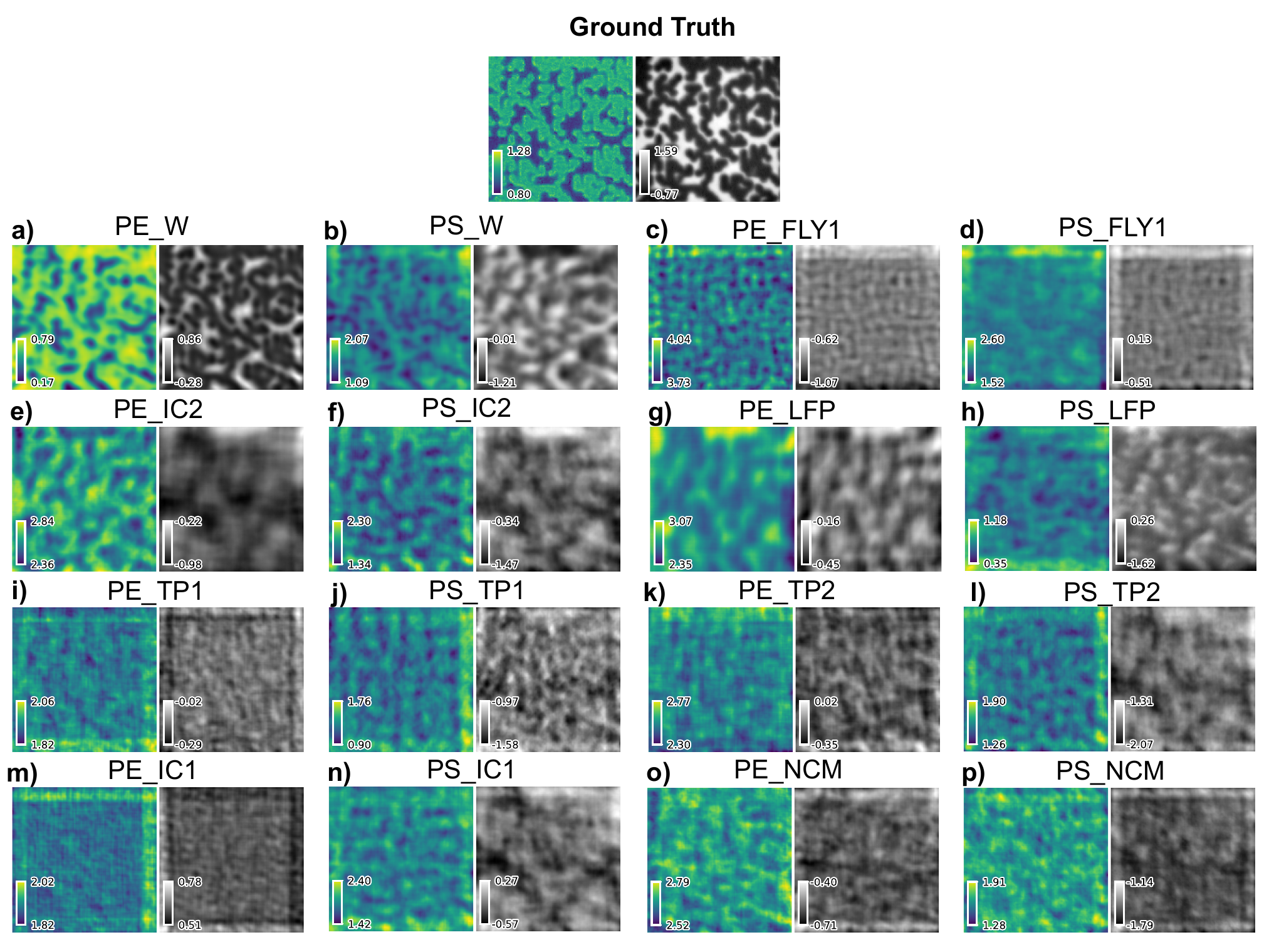}
    \caption{Predictions on the \textit{W} dataset from: a) PE\_W, b) PS\_W, c) PE\_FLY1, d) PS\_FLY1, e) PE\_IC2, f) PS\_IC2, g) PE\_LFP, h) PS\_LFP, i) PE\_TP1, j) PS\_TP1, k) PE\_TP2, l) PS\_TP2, m) PE\_IC1 n) PS\_IC1, o) PE\_NCM, p) PS\_NCM}
    \label{fig:single_dataset_transfer_cnm}
\end{figure}

\begin{figure}[htbp]
    \centering
    \includegraphics[width=0.999\textwidth]{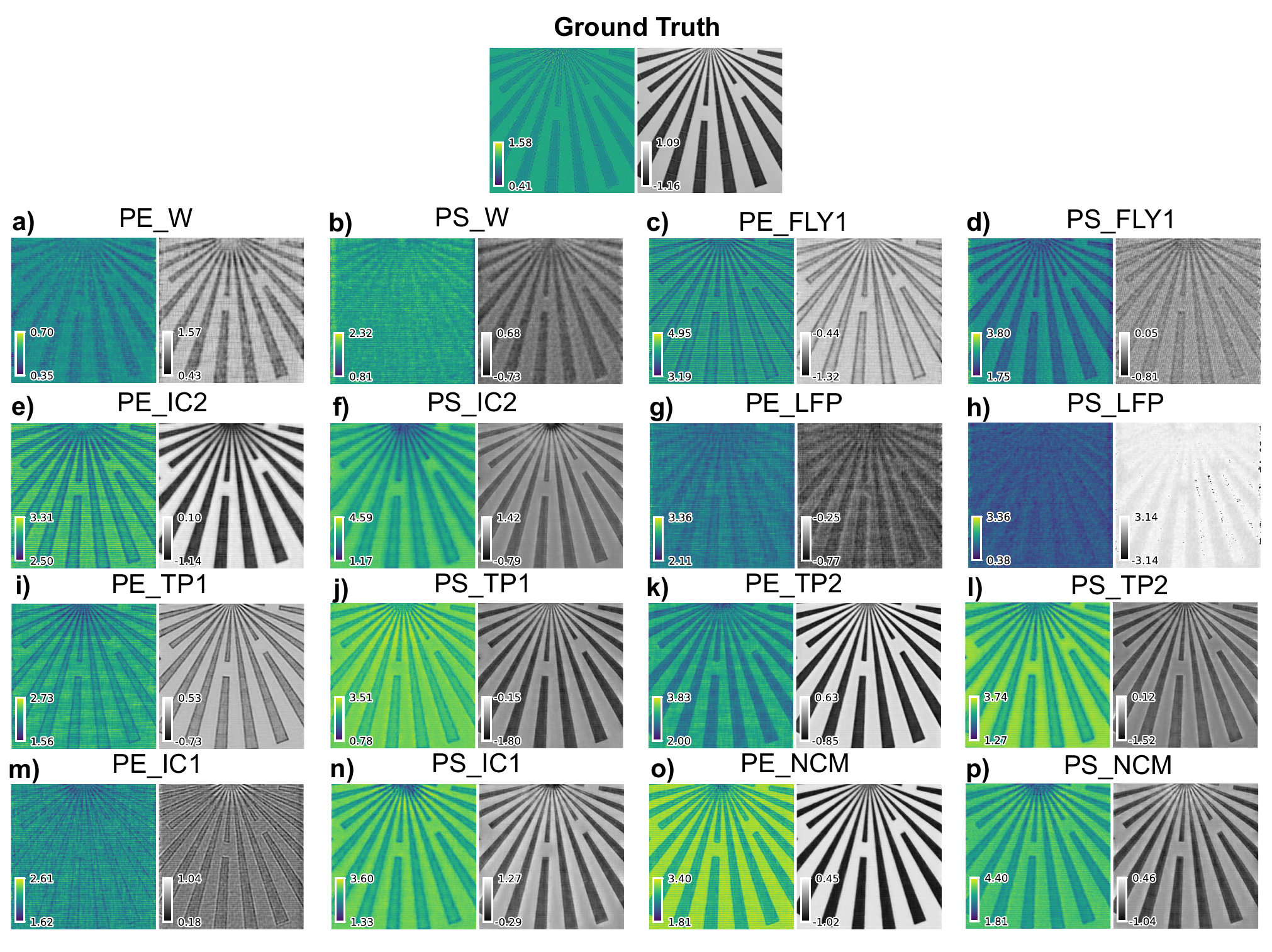}
    \caption{Predictions on the \textit{TP2} dataset from: a) PE\_W, b) PS\_W, c) PE\_FLY1, d) PS\_FLY1, e) PE\_IC2, f) PS\_IC2, g) PE\_LFP, h) PS\_LFP, i) PE\_TP1, j) PS\_TP1, k) PE\_TP2, l) PS\_TP2, m) PE\_IC1 n) PS\_IC1, o) PE\_NCM, p) PS\_NCM}
    \label{fig:single_dataset_transfer_tp2}
\end{figure}

Figure \ref{fig:single_dataset_summary} contains all FRC-AUC values from these reconstructions. In figures S\ref{fig:single_dataset_transfer_cnm}-S\ref{fig:single_dataset_transfer_tp2} we have included images of all of the model predictions (both synthetic and experiment-only) on our datasets. As stated in the main text, FRC-AUC is calculated from the integral of the FRC curve from zero frequency to the 50\% threshold (Nyquist frequency). Similar to our conclusion in the main manuscript, we can see that the similarity of test probe to training probe is a strong predictor of reconstruction quality. The naming format of models follows the main text: \textit{PE\_Experiment\_name} describes a model trained on experimental-only data from Experiment\_name, while \textit{PS\_Experiment\_name} describes a model trained on synthetic-only data generated using the probe from \textit{P\_Experiment\_name}. Dead leaves was used as the synthetic object of choice, primarily due to its precedent in the CNN literature \cite{baradad2021learning}. 

\begin{figure}[htbp]
    \centering
    \includegraphics[width=0.999\textwidth]{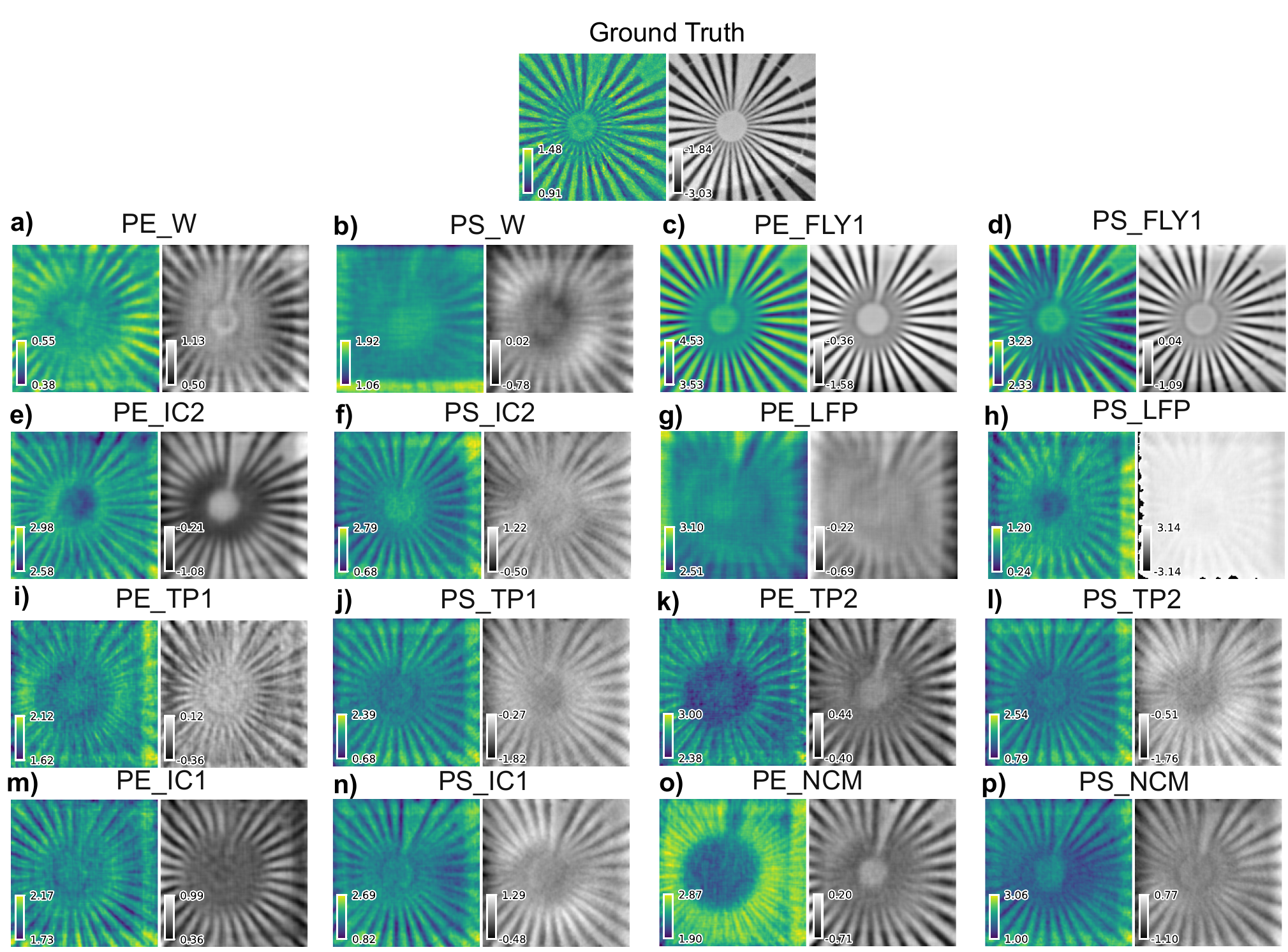}
    \caption{Predictions on the \textit{FLY1} dataset from: a) PE\_W, b) PS\_W, c) PE\_FLY1, d) PS\_FLY1, e) PE\_IC2, f) PS\_IC2, g) PE\_LFP, h) PS\_LFP, i) PE\_TP1, j) PS\_TP1, k) PE\_TP2, l) PS\_TP2, m) PE\_IC1 n) PS\_IC1, o) PE\_NCM, p) PS\_NCM}
    \label{fig:single_dataset_transfer_fly1}
\end{figure}

\begin{figure}[htbp]
    \centering
    \includegraphics[width=0.999\textwidth]{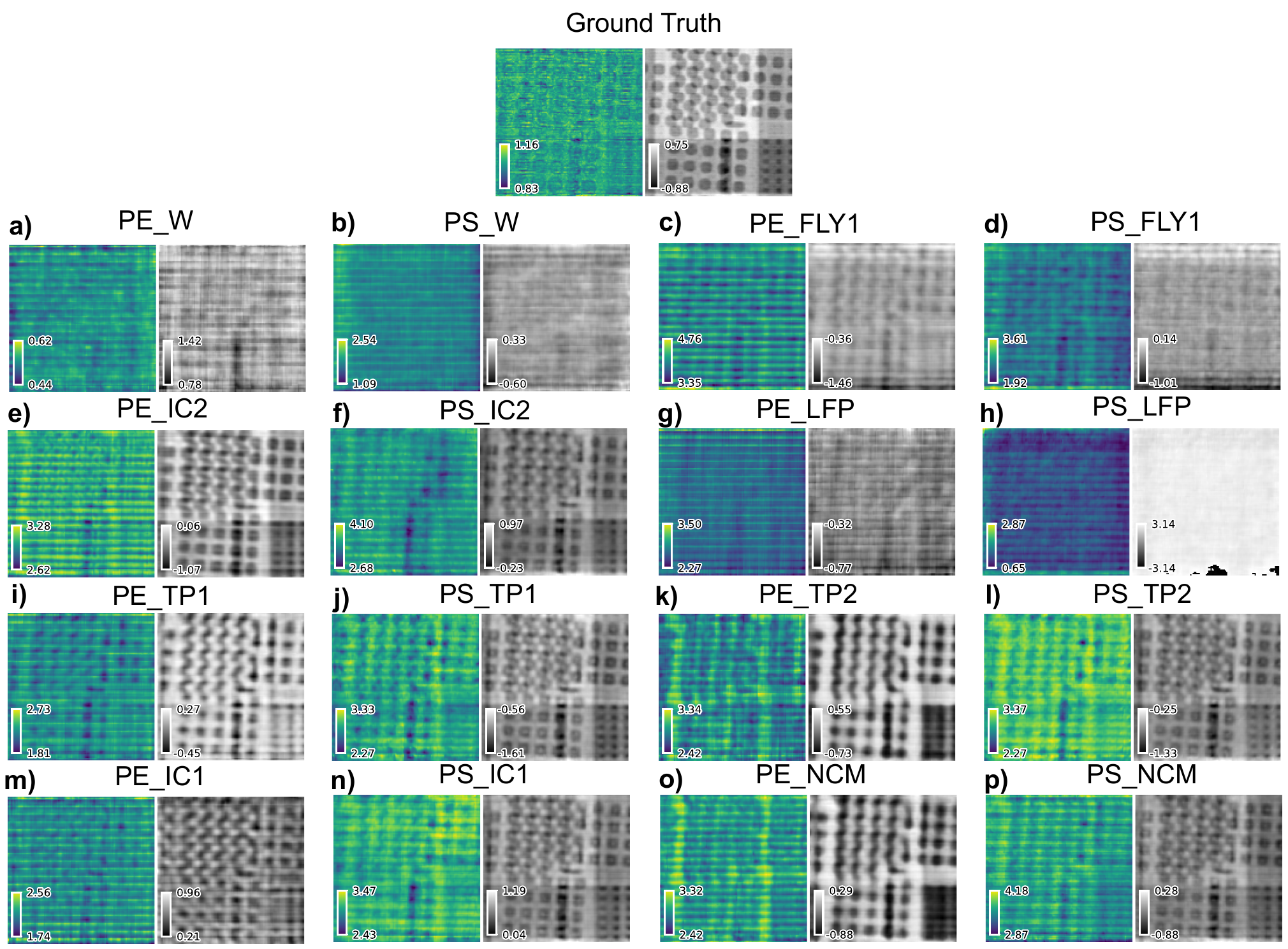}
    \caption{Predictions on the \textit{IC1} dataset from: a) PE\_W, b) PS\_W, c) PE\_FLY1, d) PS\_FLY1, e) PE\_IC2, f) PS\_IC2, g) PE\_LFP, h) PS\_LFP, i) PE\_TP1, j) PS\_TP1, k) PE\_TP2, l) PS\_TP2, m) PE\_IC1 n) PS\_IC1, o) PE\_NCM, p) PS\_NCM}
    \label{fig:single_dataset_transfer_ic1}
\end{figure}

\begin{figure}[htbp]
    \centering
    \includegraphics[width=0.999\textwidth]{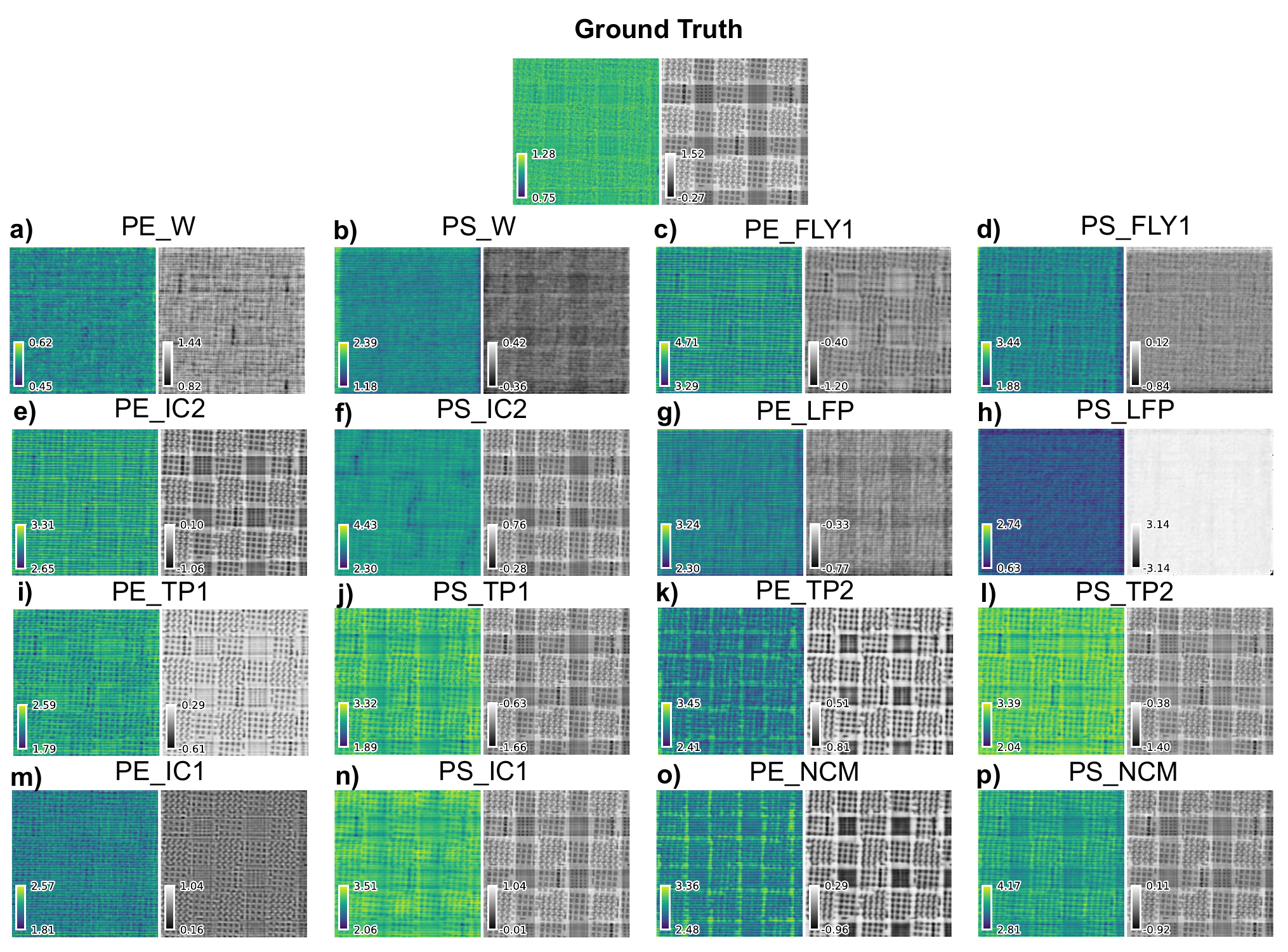}
    \caption{Predictions on the \textit{IC2} dataset from: a) PE\_W, b) PS\_W, c) PE\_FLY1, d) PS\_FLY1, e) PE\_IC2, f) PS\_IC2, g) PE\_LFP, h) PS\_LFP, i) PE\_TP1, j) PS\_TP1, k) PE\_TP2, l) PS\_TP2, m) PE\_IC1 n) PS\_IC1, o) PE\_NCM, p) PS\_NCM}
    \label{fig:single_dataset_transfer_ic2}
\end{figure}

\begin{figure}[htbp]
    \centering
    \includegraphics[width=0.999\textwidth]{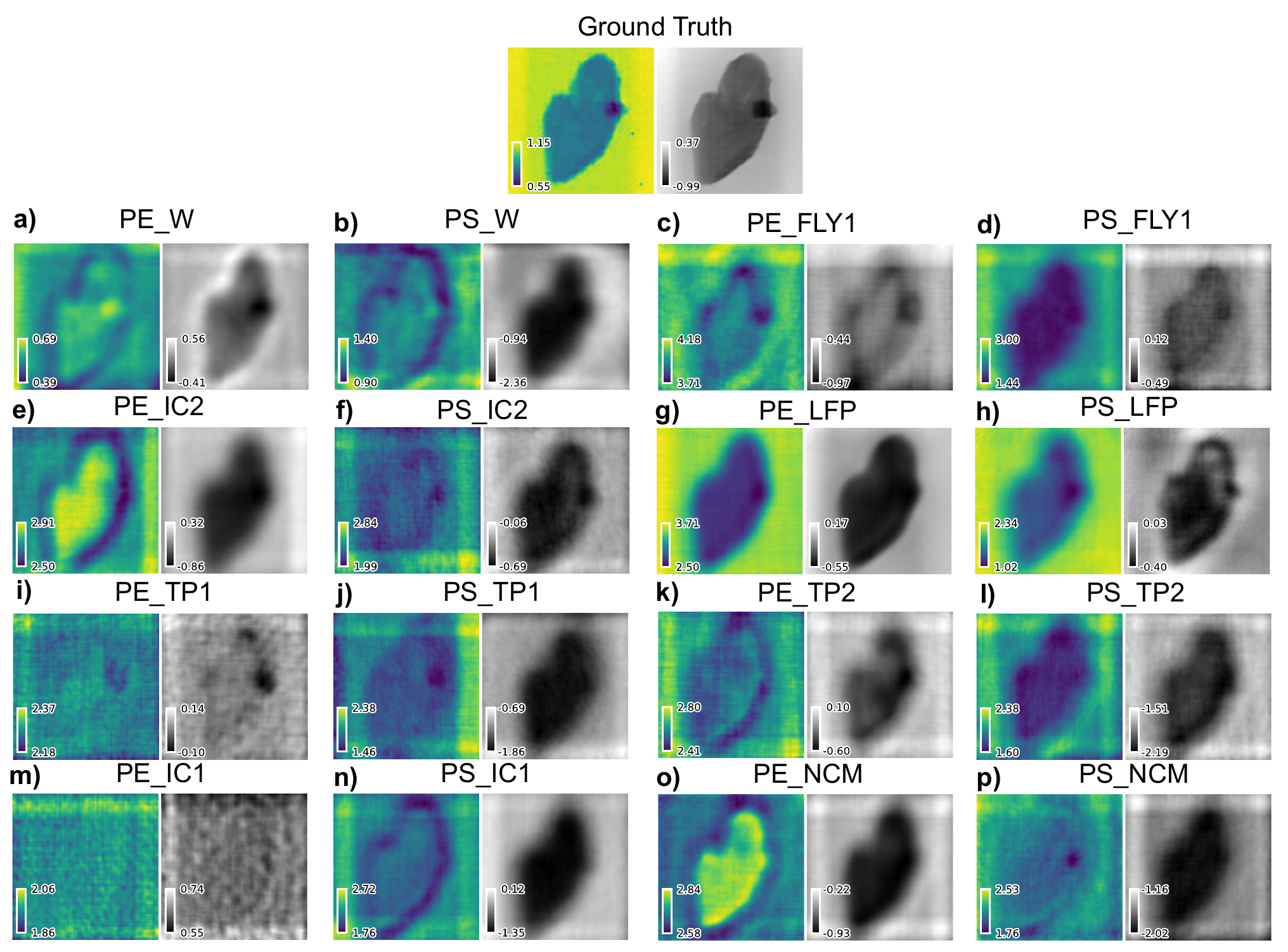}
    \caption{Predictions on the \textit{LFP} dataset from: a) PE\_W, b) PS\_W, c) PE\_FLY1, d) PS\_FLY1, e) PE\_IC2, f) PS\_IC2, g) PE\_LFP, h) PS\_LFP, i) PE\_TP1, j) PS\_TP1, k) PE\_TP2, l) PS\_TP2, m) PE\_IC1 n) PS\_IC1, o) PE\_NCM, p) PS\_NCM}
    \label{fig:single_dataset_transfer_lfp}
\end{figure}

\begin{figure}[htbp]
    \centering
    \includegraphics[width=0.999\textwidth]{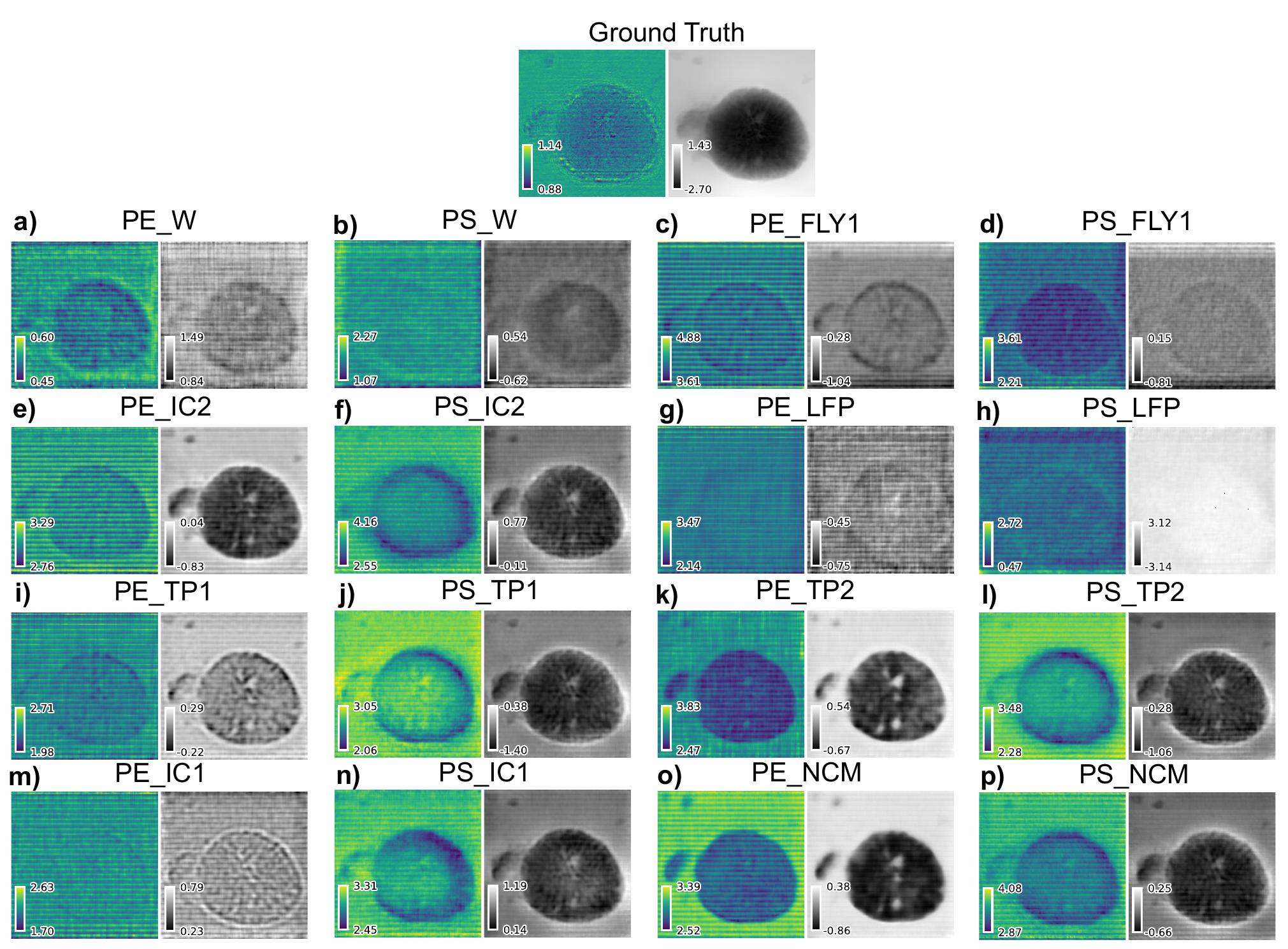}
    \caption{Predictions on the \textit{NCM} dataset from: a) PE\_W, b) PS\_W, c) PE\_FLY1, d) PS\_FLY1, e) PE\_IC2, f) PS\_IC2, g) PE\_LFP, h) PS\_LFP, i) PE\_TP1, j) PS\_TP1, k) PE\_TP2, l) PS\_TP2, m) PE\_IC1 n) PS\_IC1, o) PE\_NCM, p) PS\_NCM}
    \label{fig:single_dataset_transfer_ncm}
\end{figure}

\begin{figure}[htbp]
    \centering
    \includegraphics[width=0.999\textwidth]{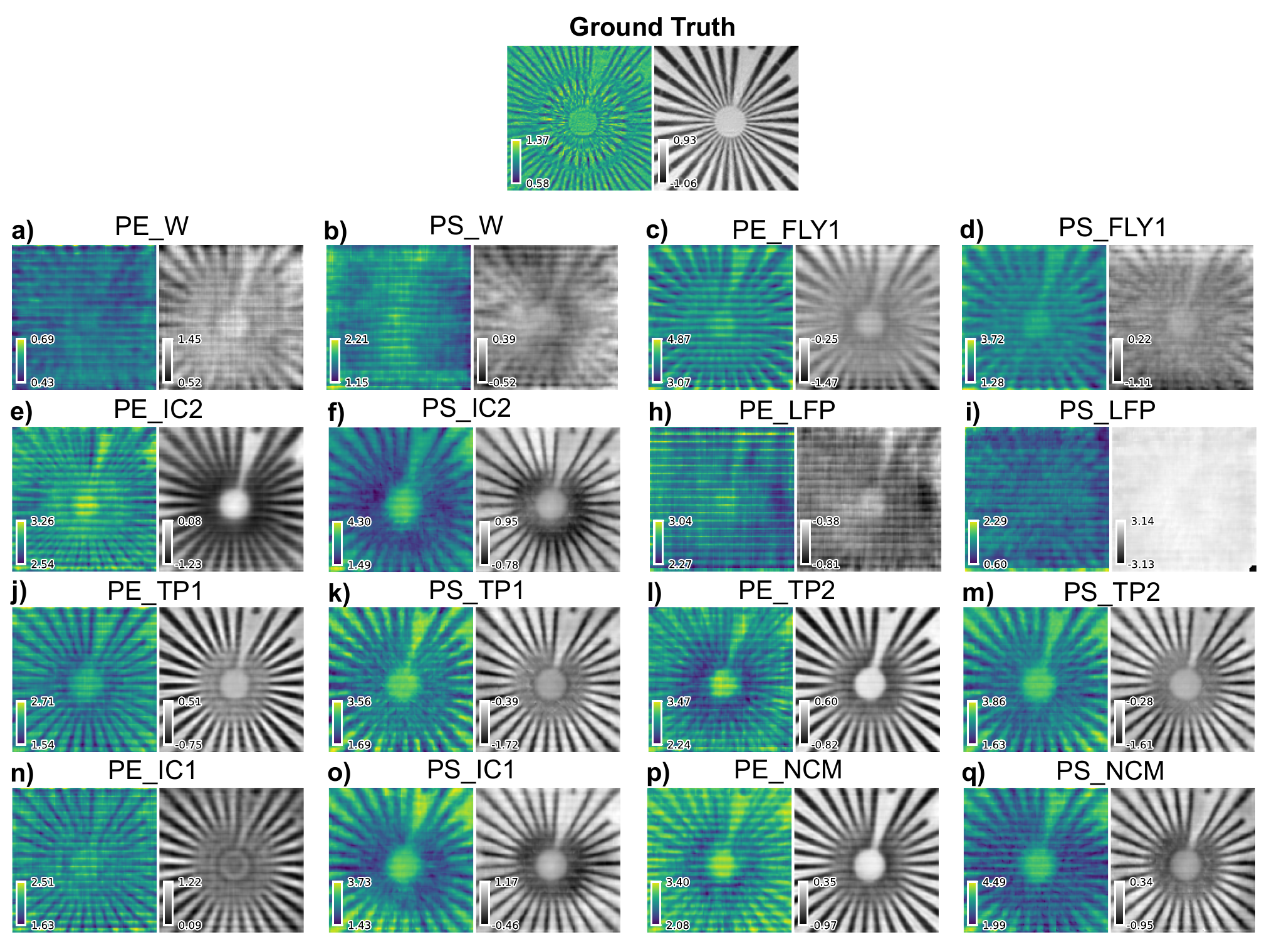}
    \caption{Predictions on the \textit{TP1} dataset from: a) PE\_W, b) PS\_W, c) PE\_FLY1, d) PS\_FLY1, e) PE\_IC2, f) PS\_IC2, g) PE\_LFP, h) PS\_LFP, i) PE\_TP1, j) PS\_TP1, k) PE\_TP2, l) PS\_TP2, m) PE\_IC1 n) PS\_IC1, o) PE\_NCM, p) PS\_NCM}
    \label{fig:single_dataset_transfer_tp1}
\end{figure}

\newpage

\subsection{LCLS data reconstruction}
We also trained a model on a dataset measured at the Linac Coherent Light Source on the X-ray pump probe instrument. There are several idiosyncracies in the dataset that make it unsuitable for the transfer learning studies shown above: (1) The LCLS pre-processing involves entire frame removals due to fluctuating illumination intensities, leading to regions with minimal overlap in the scan grid. This makes it difficult for PtychoPINN-torch to learn a meaningful representation with overlap-based inductive biases. (2) The diffraction pattern center shifts across the measurement, leading to worse reconstruction quality overall.

We were able to train an adequate model using a CDI approach where the overlap-based modules were removed from PtychoPINN-torch entirely. We show below that the reconstruction quality for a synthetically trained model, even without overlaps, is higher quality than the corresponding model trained on the raw experimental data itself. 

\begin{figure}[htbp]
    \centering
    \includegraphics[width=0.999\textwidth]{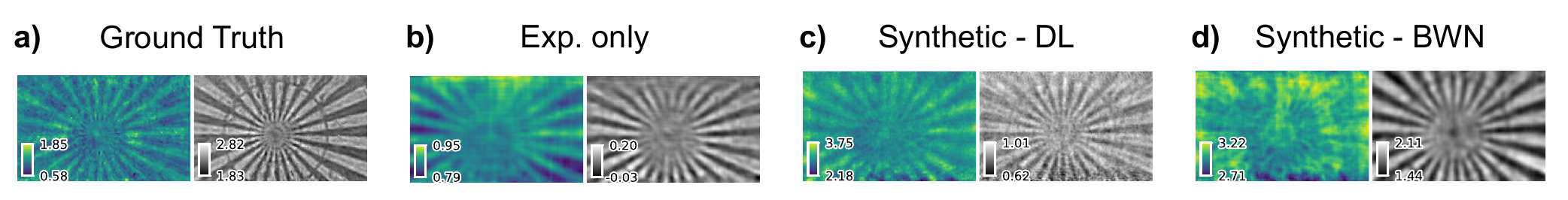}
    \caption{Predictions on the \textit{TP-LCLS dataset}. a) Ground Truth. b) Experimentally-trained model. c) Synthetic model trained on dead leaves objects. d) Synthetic model trained on blurred white noise objects.}
    \label{fig:single_dataset_transfer_lcls}
\end{figure}

\subsection{Synthetic Image Scaling}

We investigate how reconstruction quality scales with synthetic training dataset size. Based off our datasets, we observe quality saturation at around 28,000 diffraction images, with marginal improvements at larger training set sizes. Due to the combinatorial nature of evaluating a large number of models and datasets, we train all synthetic models reported in the manuscript on 28,000 images. 

The specific number of 28,000 is derived from 4 separately generated synthetic images with 7000 diffraction patterns each. This number in turn comes from sampling a 300 x 300 pixel image with pixel spacings similar to those found in experimental datasets, so PtychoPINN is able to internalize some of the spatial inter-dependencies between adjacent overlapping images.

\begin{figure}[htbp]
    \centering
    \includegraphics[width=0.999\textwidth]{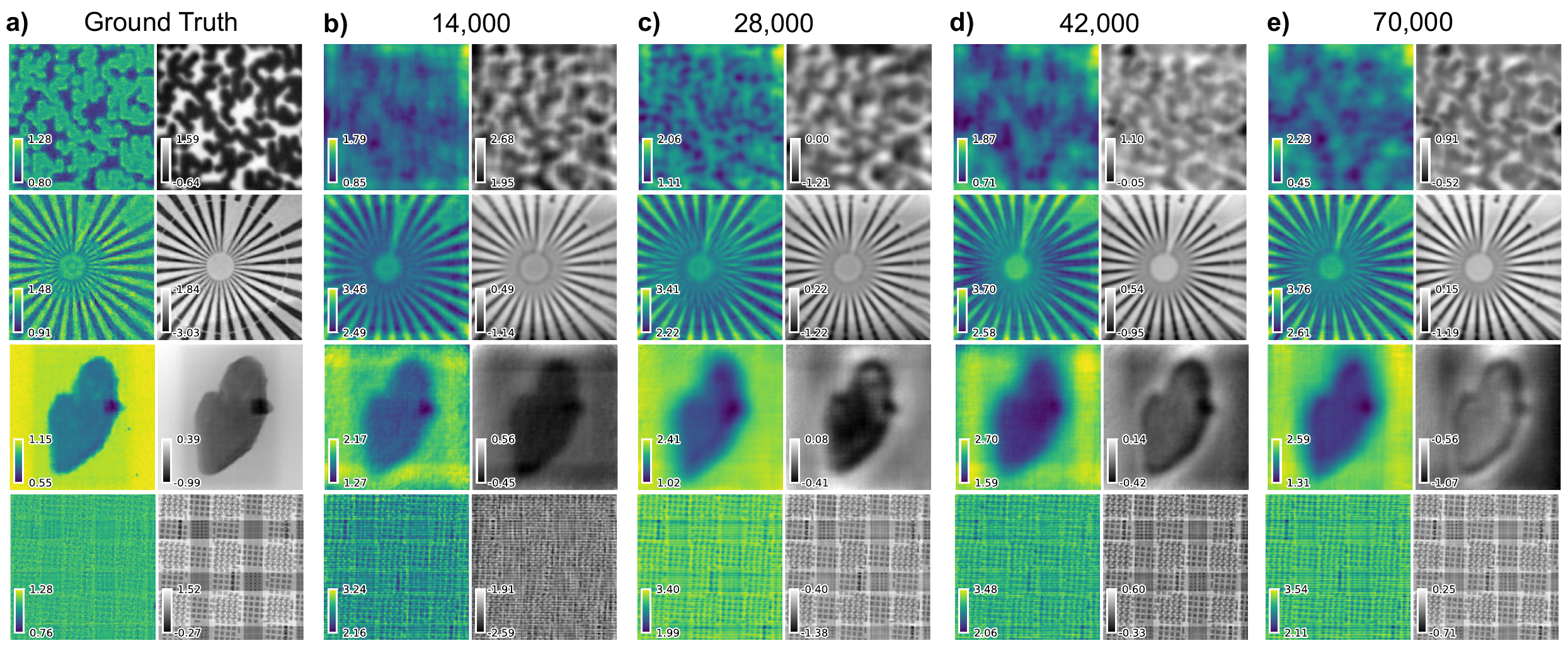}
    \caption{Exemplar reconstructions of several experiments at different synthetic training dataset sizes.}
    \label{fig:image_scaling}
\end{figure}

\section{Synthetic Objects}

\begin{figure}[htbp]
    \centering
    \includegraphics[width=0.999\textwidth]{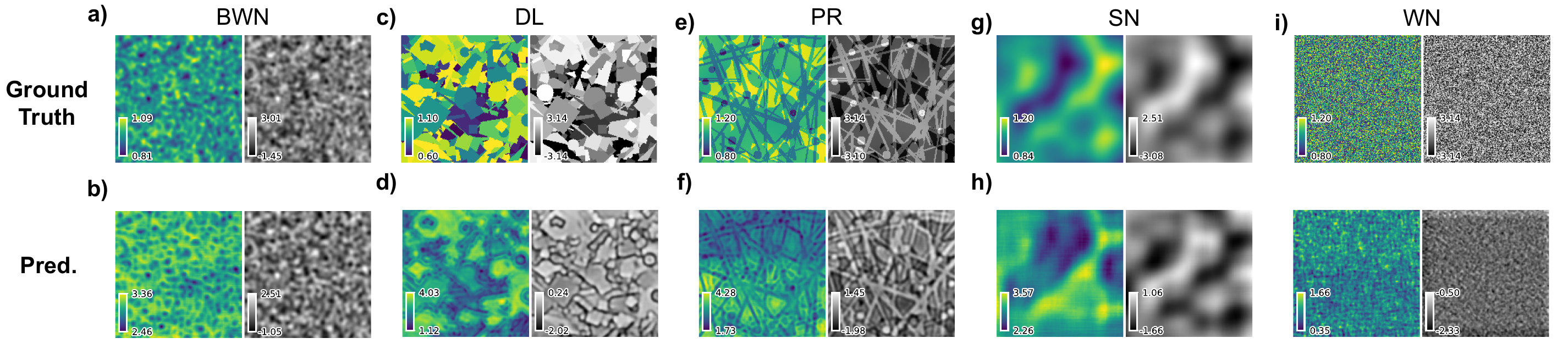}
    \caption{PtychoPINN-torch reconstructions of blurred white noise (BWN), dead leaves (DL), procedural (PR) and simplex noise (SN) via the FLY1 probe. a) BWN Ground Truth b)BWN Prediction c) DL Ground Truth d) DL Prediction e) PR Ground Truth f) PR Prediction} g) SN Ground Truth h) SN Prediction
    \label{fig:training_object_predictions}
\end{figure}

\begin{figure}[htbp]
    \centering
    \includegraphics[width=0.999\textwidth]{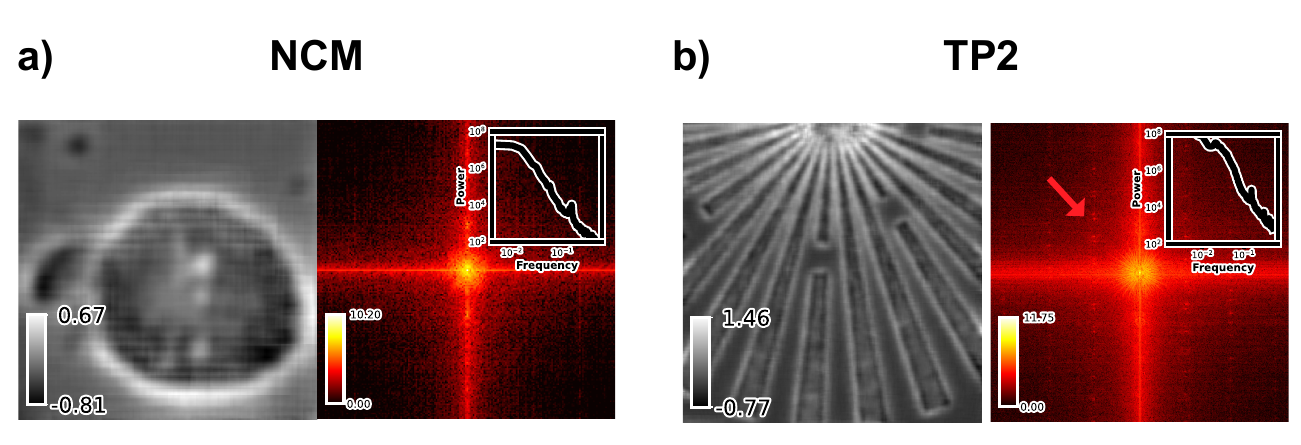}
    \caption{Simplex noise reconstructions of NCM and TP2. TP2-SN contains sharp line features absent in SN, and NCM-SN exhibits circular features smaller than SN's maximum characteristic length scale. }
    \label{fig:training_object_sn}
\end{figure}

\newpage

\section{Dataset Power Spectral Densities}

\begin{figure}[htbp]
    \centering
    \includegraphics[width=0.999\textwidth]{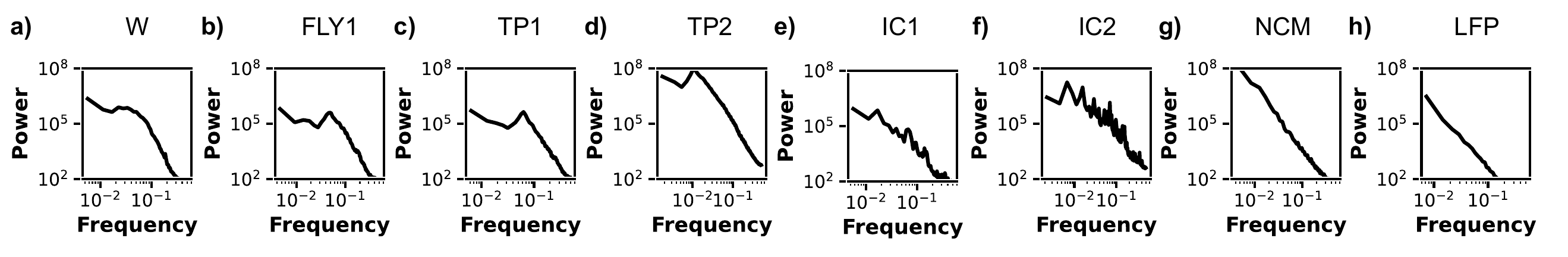}
    \caption{1D Power Spectral Densities for all experimental datasets: a) W, b) FLY1, c) TP1, d) TP2, e) IC1, f) IC2, g) NCM, h) LFP}
    \label{fig:dataset_psds}
\end{figure}

\section{Benchmarking Inference Speed}

Here we provide full benchmarking results on a machine with an Intel Xeon 3.9 GhZ CPU with 64 GB RAM and 2 Nvidia RTX A4500 in Table \ref{tab:timing}. For DNN inference, we executed the same inference code on each dataset 5 times, taking the average and standard deviation of different steps in the process. Iterative reconstruction using pty-chi was executed once, as the time per iteration averages out over 500 total iterations. All other reconstruction algorithms and settings are the same as those describes previously, except for iteration number.

\label{tab:timing}
\vspace{12pt}
\begin{tabular}{@{}ccccccc@{}}
\toprule
\textbf{Name} & \textbf{\# of Images} & \textbf{\shortstack{Pty-Chi \\ 500 iter. (s)}} & \textbf{Inference (s)} & \textbf{Assembly (s)} & \textbf{Speedup Ratio} \\
\midrule
\textit{IC1} & 1,443 & 6.4 & $0.52 \pm 0 $& $0.03 \pm 0.01$ & 11.7$\times$ \\
\textit{TP1} & 1,443 & 60.1 & $0.52 \pm 0$ & $0.04 \pm 0.04$ & 107.3$\times$ \\
\textit{NCM} & 2,466 & 128.5 & $0.61 \pm 0$ & $0.03 \pm 0.01$ & 200.7$\times$ \\
\textit{LFP} & 5,625 & 191.4 & $0.82 \pm 0 $& $0.04 \pm 0.03$ & 222.5$\times$ \\
\textit{TP2} & 7,709 & 376.0 & $0.91 \pm 0$ & $0.02 \pm 0$ & 404.3$\times$
\\
\textit{FLY1} & 10,304 & 453.6 & $1.06 \pm 0$ & $0.07 \pm 0$ & 401$\times$ \\
\textit{IC2} & 9,466 & 469.8 & $1.04 \pm 0.01$ & $0.03 \pm 0.01$ & 439.1$\times$ \\
\textit{W} & 25,921 & 1088.1 & $1.91 \pm 0.01$ & $0.05 \pm 0$ & 555 $\times$ \\
\bottomrule
\end{tabular}

\section{Supplemental References}


\begin{enumerate}
\item Baradad Jurjo, M., Wulff, J., Wang, T., Isola, P. \& Torralba, A. Learning to see by looking
at noise. \textit{Advances in Neural Information Processing Systems} \textbf{34} 2556-2569 (2021).
\end{enumerate}

\end{document}